\documentclass[preprint,11pt]{elsarticle}
\usepackage{lineno,hyperref}
\usepackage[ruled,vlined]{algorithm2e}
\usepackage{algorithmic,algorithm2e,float}
\usepackage{graphicx}
\usepackage{caption}
\usepackage{amsmath, mathtools}
\usepackage{natbib}
\biboptions{sort&compress}
\usepackage{etoolbox}
\usepackage{longtable}
\usepackage{mathtools}
\usepackage[mathscr]{eucal}
\usepackage{amssymb}
\usepackage{adjustbox}
\usepackage{amsmath} 

\journal{Eng Appl Artif Intel}
\begin{document}

\begin{frontmatter}
\title{\textbf{Adaptable Automation with Modular Deep Reinforcement Learning and Policy Transfer}}

\author{Zohreh Raziei, Mohsen Moghaddam\corref{myfootnote}}
\address{Department of Mechanical and Industrial Engineering, Northeastern University, Boston, MA 02115, United States}
\cortext[myfootnote]{Corresponding Author \\ raziei.z@northeastern.edu, mohsen@northeastern.edu (Zohreh Raziei, Mohsen Moghaddam)}


\begin{abstract}

Recent advances in deep Reinforcement Learning (RL) have created unprecedented opportunities for intelligent automation, where a machine can autonomously learn an optimal policy for performing a given task. However, current deep RL algorithms predominantly specialize in a narrow range of tasks, are sample inefficient, and lack sufficient stability, which in turn hinder their industrial adoption. This article tackles this limitation by developing and testing a Hyper-Actor Soft Actor-Critic (HASAC) RL framework based on the notions of task modularization and transfer learning. The goal of the proposed HASAC is to enhance the adaptability of an agent to new tasks by transferring the learned policies of former tasks to the new task via a ``hyper-actor''. The HASAC framework is tested on a new virtual robotic manipulation benchmark, Meta-World. Numerical experiments show superior performance by HASAC over state-of-the-art deep RL algorithms in terms of reward value, success rate, and task completion time.
\end{abstract}

\end{frontmatter}


\section{Introduction}
\label{introduction}

Advances in machine learning, robotics, and computing technologies are giving rise to a new generation of ``intelligent'' automation technologies that can transform various industries such as manufacturing, healthcare, defense, transportation, and agriculture, among others. Automation and robotics are a central building-block of the above industries---from processing and part transfer robots in factories to assistive and patient monitoring robots in hospitals, unmanned aerial drones/vehicles in agricultural/defense fields, and mobile robots in warehouses and fulfillment centers \cite{koren1999reconfigurable,trentesaux2009distributed,leitao2012bio,moghaddam2018reference, yu2019reinforcement, ruiz2020predictive}. The need for ``intelligence'' in such automation systems stems from the fact that most robotic operations in industry are currently limited to rote and repetitive tasks performed within structured environments. This leaves an entire swath of more complex tasks with high degrees of uncertainty and dynamic environments \cite{weyer2015towards} difficult or even impossible to automate. Examples include maintenance and material handling for producing the desired product in manufacturing systems \cite{arinez2020artificial}, robot surgeries and pharmacy automation in healthcare systems \cite{bhattacharya2016review}, safe working environments in disaster management for deep-sea operation, and nuclear energy \cite{robert2017growing}, fruit picking, crop sensing, and selective weeding in agriculture systems \cite{marinoudi2019robotics} .

A fundamental question concerning the notion of intelligent automation in this context then becomes: How can we enable adaptable industrial automation systems that can analyze and act upon their perceived environment rather than merely executing a set of predefined programs? Adaptability is among the key characteristics of industrial automation systems in response to unpredictable changes or disruptions in the process \cite{ivanov2020digital}. Recent developments in deep learning techniques along with the emergence of collaborative robots (cobots) as flexible and adaptive automation ``tools'' \cite{chou2018fourth} have enabled the preliminaries for incorporating intelligence and learning capabilities into the current fixed, repetitive, task-oriented industrial manipulators \cite{pane2019reinforcement}. The challenge, however, is to design efficient control algorithms and architectures that enable the robots to modify their behavior to cope with uncertain situations and automatically improve their performance over time, with minimal need for reprogramming or manual tuning of the robot's behavior.

Reinforcement Learning (RL) has achieved tremendous success in recent years in enabling intelligent control of robotic manipulators in simulated environments or physical lab settings \cite{stone2005reinforcement, riedmiller2007learning}. RL allows agents to learn behaviors through interaction with their environment and generalize the learned policies to new unseen variations of the same task without changing the model itself \cite{sutton2018reinforcement}. Yet, current studies are predominantly focused on simplified tasks that cannot be scaled up to challenging, real-life problems \cite{bansal2017emergent, heess2017emergence}. This is caused by a number of major limitations of current RL algorithms for continuous control which we refers some of them below:

\begin{itemize}
\setlength\itemsep{0.1em}
    \item \textit{Task specialization}. Deep RL algorithms often require reprogramming and relearning on new tasks from scratch. This limitation is due to the lack of efficient mechanisms for transfer of learned policies between tasks. For instance, a cobot can only move objects of similar shape and size in a defined environment or assemble a particular set of chips on a board. Task specialization is one of the critical gaps in RL for continuous control, which has been extensively investigated by many researchers in recent years \cite{levine2016end,duan1611rl2,tamar2017learning,yu2019meta,wang2016learning, xu2018meta, battaglia2018relational, botvinick2019reinforcement,lebensold2019actor, humplik2019meta, dulac2020empirical}.
    
    \item \textit{Sample inefficiency}. Deep RL algorithms often require massive training datasets \cite{wang2016learning,botvinick2019reinforcement}, even for learning tasks that are relatively simpler than complex, real-life tasks; e.g., playing Atari games \cite{mnih2015human}, Go \cite{silver2017alphago}, and poker \cite{moravvcik2017deepstack}. However, there is still the problem of handling multi-tasks systems. The other critical challenge for the RL algorithms is the problem of stability - especially in high-dimensional continuous action spaces \cite{he2007reinforcement} - and being sensitive to hyperparameter settings \cite{duan2016benchmarking}.
    
    \item \textit{Stability}. Hyperparameters must be set carefully for different problem settings to obtain good results. However, most deep RL algorithms, such as Deep Deterministic Policy Gradient (DDPG) \cite{lillicrap2015continuous} and Twin Delayed Deep Deterministic Policy Gradient (TD3PG) \cite{kim2020motion} are too sensitive to hyperparameter tuning \cite{duan1611rl2}. In those algorithms, a minor variation in hyperparameters can lead to completely different performance and results. Haarnoja et al. \cite{haarnoja2018soft} propose an off-policy maximum entropy actor-critic-based method, Soft Actor-Critic (SAC), that addresses both sample efficiency and stability issues. Nevertheless, the problem of balancing between exploitation and exploration still remains to be addressed.  
    \item \textit{Inductive bias}. Training an appropriate model requires training with a large number of samples. Meta RL tackles this limitation by training the agent on a series of interrelated tasks drawn from a given distribution or a prior \cite{duan1611rl2, zambaldi2018deep, gupta2018unsupervised, botvinick2019reinforcement, jabri2019unsupervised} to leverage the agent's inductive bias for learning new tasks rapidly or adapting to new environments.
\end{itemize}

Motivated by the aforementioned limitations, this article aims at building and testing a new deep RL framework, based on the actor-critic\cite{haarnoja2018soft}, to enable a robot to efficiently adapt to new variations of its previously learned tasks by sharing parameters across tasks with both parametric and non-parametric variations. The proposed framework is built upon the notions of \textit{task modularization} \cite{alet2018modular} and \textit{Transfer Learning} (TL) \cite{taylor2009transfer}, and the idea of training a general neural network at both module and task levels. The underlying idea is that transferring the learned knowledge across different interrelated tasks can potentially alleviate the problems of task specialization through multi-task learning \cite{zhang2017survey, rahmatizadeh2018vision, fox2019multi} and sample efficiency through parameter sharing between tasks and/or task modules \cite{pinto2017learning}. The proposed framework contributes the following features to realize the aforementioned goal of the study:

\begin{itemize}
\setlength\itemsep{0.1em}
    \item \textit{Task modularization}. In complex industrial applications of robots, adaptability can be achieved by modularizing the learning tasks, learning optimal policies for each individual module, and then transferring learned policies across modules of different interrelated tasks with both parametric and non-parametric variations. Therefore, we build on the notion of task modularity for faster learning of a series of manipulation tasks. Using the reward function of each task, smaller modules of the task can be obtained. These modules are also used in order to enhance the training of the task by training their modules and feeding them to the task actor in proposed actor-critic architecture. 
    \item \textit{Parameter sharing}. In order to increase learning speed and decrease memory usage, we propose the idea of hyper-networks (Hyper-Actor) to automatically transfer the knowledge between the modules of the task and between a sequence of interrelated tasks during training with modifying SAC algorithms.
    \item \textit{Multi-task learning}. We generalize the idea of our modularized training to the multi-task level. Hence, two types of hyper-networks trained on all tasks and modules simultaneously. This forms the meta-learning structure of our actor-critic design architecture by obtaining a multi-task modularized trained actor-network, that can be used for testing a completely new task.
    \item \textit{Exploration-exploitation trade-off}. We define and employ a new replay buffer in addition to the existing one to enable the SAC algorithm to achieve a stable trade-off between training the neural networks and memory requirement, and faster training.
\end{itemize}

This article is to propose a deep neural network architecture for transferring knowledge to test a new task from different trained tasks with employing  based methods on one of the recent robotics benchmark named \textit{Meta-World} \cite{yu2019meta}. This benchmark contains different robotic manipulation tasks for a single task and multi-task learning. However, the critical issue is how to adapt the state space and reward function of tasks defined in the Meat-World with the proposed algorithms because Meta-World, by default, has a sparse reward function for each task manipulation based on the distance. 

Section \ref{background} provides an overview of background and related work in the areas of transfer leaning, actor-critic-based RL, and meta learning. Section \ref{preliminaries} presents the theoretical and modeling preliminaries. Section \ref{framework} discusses the proposed framework in detail. Section \ref{experiments} presents the experimental results and analyses. Section \ref{conclusions} provides conclusions and a summary of limitations and directions for future research.

\section{Background and Related Work}
\label{background}

RL algorithms attempt to optimize the behavior of an agent to solve a given task. The agent interacts with the environment by taking an action. Any phenomenon that is beyond the control of the agent to manipulate is considered as a part of the environment \cite{sutton2018reinforcement}. The limited scalability of RL algorithms to problems with large state and action spaces, due to the so-called ``curse of dimensionality'', has hindered the implementation of RL algorithms to complex, industrial problems \cite{gueant2019deep}. In recent years, deep RL achieved incredible success by reaching super-human performance in learning how to play Atari games \cite{mnih2015human}. The results lead to the development of more challenging tasks by DeepMind such as AlphaGo \cite{silver2016mastering} in 2016 and AlphaStar \cite{vinyals2019alphastar} in 2019. In 2020, Garc{\'\i}a and Shafie \cite{garcia2020teaching} proposed a safe RL algorithm to teach humanoid robots how to walk faster. 

\subsection{Transfer Learning}
In the context of RL, transfer learning can enable the transfer of prior knowledge from learning an old task to a new task to help the agent master the new tasks faster \cite{taylor2009transfer}. It helps decrease the overall time for learning a complex task. The fundamental idea of transfer learning is to create lifelong machine learning methods that reuse previously learned knowledge in new learning environments for higher task efficiently \cite{pan2009survey}. This is not a new idea and has been studied for decades \cite{skinner1965science}. To successfully realize the vision of transfer learning, an RL agent needs to take following steps \cite{taylor2009transfer}:

\begin{itemize}
\setlength\itemsep{0.07em}
    \item Select a set of \textit{source tasks} from which to transfer to a \textit{target task}.
    \item Learn the relationship between the source task(s) and the target task.
    \item Transfer knowledge from the source tasks to the target task.
\end{itemize}
Transfer learning helps augment the efficiency of deep RL methods for low-dimensional data representations. It also improves the learning rate when there is a combination of different methods \cite{celiberto2016transfer}. Moreover, it helps with transferring information between different modules, tasks, or robots \cite{barrett2010transfer}. 

Parisotto et al. \cite{parisotto2015actor} define a novel multi-task and transfer learning algorithm, actor-mimic, for learning multiple tasks simultaneously. This method trains a single policy network by implementing deep RL and model compression techniques for learning a set of distinct tasks. Devin et al. \cite{devin2017learning} propose a neural network architecture for decomposing policy into task-specific and robot-specific modules for sharing tasks and robot information. They present the effectiveness of their transfer method for zero-shot generalization with different robots and tasks. 

To measure the improvement when transfer, the following evaluation metrics can be use: \textit{jumpstart}, \textit{asymptotic performance}, \textit{total reward}, \textit{time to threshold}, and \textit{transfer ratio} \cite{taylor2007cross}. The first four metrics are appropriate for fully autonomous scenarios, because the time spent for learning the source tasks is not taken into account. 

\subsection{Actor-Critic Based Algorithms}

In RL algorithms, it is essential to learn value function in addition to the policy, because the policy will be updated by the information received from the value function. This procedure is the core idea behind the actor-critic method \cite{konda2000actor, wiering2008ensemble, sanchez2015priori, lawhead2019bounded}. Actor-critic consists of two models, an actor and a critic. The critic approximates the action-value function or state-value function, which is then used to update the actor’s policy parameters \cite{konda2000actor}. The number of parameters that the actor needs to update is small compared to the number of states. Therefore, a projection of the value function is computed onto a low-dimensional subspace spanned by a set of basis functions. It is demonstrated by parameterization of the actor. 

Some actor-critic algorithms update the actor through on-policy gradient formulation. Same as off-policy methods, the goal of on-policy training is to improve stability. However, it is proven that off-policy algorithms achieve relatively higher stability and sample efficiency. 
Similar to the actor-critic model, other off-policy algorithms reuse the experience for updating the policy \cite{degris2012off}. One of the extensions of the original actor-critic method is defined based on the difference between the value function and a baseline value, which is known as Advantage Actor-Critic (A2C) method \cite{mnih2016asynchronous}. Deep Deterministic Policy Gradient (DDPG) \cite{lillicrap2015continuous}, the deep learning alternative of the deterministic policy gradient with function approximation \cite{silver2014deterministic}, is one of the most popular off-policy actor-critic algorithms. It applies a $Q$-function estimator to allow off-policy learning and a deterministic actor to maximize the $Q$-function \cite{lillicrap2015continuous}. 
The interaction between the $Q$-function and the deterministic actor causes difficulties in stabilizing DDPG for hyperparameter settings \cite{duan2016benchmarking}. The main limitation of the DDPG algorithm is the oscillations in performance in unstable environments, which hinders their use for learning complex, high-dimensional tasks. Moreover, the results presented by Popov et al. \cite{popov2017data} confirm the fact that DDPG is successful for the cases with binary reward functions. 

Mnih et al. \cite{mnih2016asynchronous} present Asynchronous Advantage Actor-Critic (A3C) algorithm for stabilizing
the variation caused by training asynchronous parallel agents with accumulated updates. Haarnoja et al. \cite{haarnoja2018soft} present an off-policy algorithm based on the maximum entropy framework. This algorithm increases the standard maximum reward of RL objective function with an entropy maximization term, named Soft Actor-Critic (SAC). SAC seeks stability, exploration, and robustness simultaneously \cite{ziebart2010modeling}. Empirical results show that SAC outperforms prior off-policy and on-policy algorithms in terms of both performance and sample efficiency.

\subsection{Task Modularization}
A key enabler for adaptability and quick response to the variations in the tasks assigned to a machine is task modularization \cite{feldmann2012modularity}. In manufacturing systems, for example, such adaptability is the main requirement for efficient transition from traditional mass-production systems to mass-customization and eventually to ``lot-size of one'', where each task (e.g., product type) is different from its predecessor and successor tasks. Modularity is a fundamental concept in architecting engineering products, processes, and organizations \cite{eppinger2015product}, and has been identified as an adaptability mechanism in different research domains \cite{simon1991architecture}. The notion of task modularity can therefore be incorporated in RL algorithms to enable the agent to adapt to different tasks with both parametric and non-parametric variations more efficiently. The history of using task modularity for increasing the adaptability of RL agents dates back to the early 90's \cite{gianetto2015network, mosleh2017fair}. 

Two general approaches to modularity in the context of AI have been proposed in literature. The first approach is based on hierarchical learning, in which most proposed approaches consist of separate mechanisms for task decomposition \cite{singh1992efficient}. To compute a joint policy, the agent combines the modules' action preference, where each module is considered as a distinct policy agent \cite{russell2003q, simpkins2019composable}. In the second approach, the notion of modularity is directly applied to the deep neural network architecture. Our proposed framework follows the second approach, which is applied by some recent studies to create more reliable RL systems \cite{andreas2016neural, chitnis2019learning, devin2017learning}. The overall idea is built on the assumption that modules are reusable neural network functions. Therefore, they can be pre-trained and recombined in different ways to tackle new tasks. Thus, instead of training a single network on a complex task that would take a long time, one can train on many different task module networks. The idea of using a set of reusable neural network modules has been applied to applications such as reasoning problems, which define a general-purpose approach for learning collections of neural modules \cite{andreas2016neural}. Robot task and motion planning is another application of modular meta-learning \cite{chitnis2019learning}. Our framework builds on the work of Andreas et al. \cite{alet2018modular}, which uses a set of neural network modules on two challenging datasets of supervised robot learning problems. In their approach, the network modules can be re-tuned and recombined for solving new tasks. We further extend this idea into deep RL applications, as described in Section \ref{framework}.

\subsection{Meta Reinforcement Learning}
Meta learning is an emerging concept in deep learning with roots in psychology, also known as ``learning to learn'' \cite{harlow1949formation}. Meta learning aims to leverage past experience to reduce learning time and increase efficiency in adapting to new tasks \cite{schweighofer2003meta, schaul2010metalearning}. Formally, the task is defined as:
\begin{equation}
\mathcal{T}=\{\mathcal{L}(\mathbf{s}_{1}, \mathbf{a}_{1}, \ldots, \mathbf{s}_{K}, \mathbf{a}_{K}), p(\mathbf{s}_{1}), p(\mathbf{s}_{t+1} | \mathbf{s}_{t}, \mathbf{a}_{t}), K\},
\end{equation}
where $\mathcal{L}$ is loss function, $p(\mathbf{s}_{1})$ is the distribution over initial state, and $p(\mathbf{s}_{t+1} | \mathbf{s}_{t}, \mathbf{a}_{t})$ is the transition distribution over episode length $K$. Therefore, the distribution over tasks is define as $p(\mathcal{T})$. 

Meta learning algorithms are based on two main assumptions. First, the meta-training and meta-test tasks are induced from the same distribution. Second, the task distribution exhibits a shared structure that can be utilized for efficient adaptation to new tasks. Meta learning has three requirements \cite{lemke2015metalearning} including learning the subsystems, dynamically choosing the learning bias, and achieving experience by exploiting meta knowledge extracted in a previous learning episode on a unique dataset. It trains the agent on a series of interrelated tasks drawn from a given distribution or a prior (\cite{duan1611rl2, gupta2018unsupervised, botvinick2019reinforcement, jabri2019unsupervised}).

One of the powerful meta learning algorithms is Model-Agnostic Meta-Learning (MAML) algorithm \cite{finn2017model}. MAML aims to find an efficient parameters initialization to achieve the best performance after one update on a series of tasks. Grant et al. \cite{grant2018recasting} propose a more accurate estimate of the original MAML by applying a Laplace approximation to the posterior distribution over task-specific parameters. Finn et al. \cite{finn2018probabilistic} present a probabilistic extension of MAML, which adapts to new tasks via gradient descent. In meta-test time, the algorithm is adapted through noise injection into gradient descent. Yu et al. \cite{yu2019meta} provide experimental results for 50 robotics environments in ``Meta-World''. The experiments were performed on various multi-task and meta-learning algorithms such as MAML, $RL^2$, and PEARL \cite{rakelly2019efficient, duan1611rl2}.

Meta learning addresses two fundamental limitations of deep learning algorithms: sample inefficiency and single-task specialization \cite{wang2016learning, botvinick2019reinforcement}. These limitations result in weak inductive bias and incremental parameter adjustment \cite{botvinick2019reinforcement}. Meta RL improves the exploration process by augmenting the policy input \cite{gupta2018meta}. Rakelly et al. \cite{rakelly2019efficient} propose a meta RL algorithm, PEARL, which is an off-policy RL algorithm that enables posterior sampling for exploration at test time. Humplik et al. \cite{humplik2019meta} propose a meta RL method for separately learning the policy and the task by using the privileged information of the unobserved task. They claim that the method enhances the performance of both on-policy and off-policy meta RL algorithms. Meta RL has recently been applied for robotic manipulation in manufacturing systems. Schoettler et al. \cite{schoettler2020meta} propose a meta RL approach to solve the complex industrial robotic insertion tasks under 20 trails. In their experiment, they perform connector assembly on a 3D-printed gear insertion task. They use MuJoCo engine \cite{todorov2012mujoco} to simulate the tasks.

\section{Preliminaries}
\label{preliminaries}

An RL problem can be modeled as a Markov Decision Process (MDP), which is a stochastic process that satisfies the Markov property \cite{sutton2018reinforcement} defined by a tuple $\langle \mathcal{S}, \mathcal{A}, p, r, \gamma \rangle$, where $\mathcal{S}$ is state space, $\mathcal{A}$ is action space, $p: \mathcal{S} \times \mathcal{A} \rightarrow \mathbb{P}_{\mathbf{s}}$ is state transition distribution, $r$ is reward function, and $\gamma$ is a discount factor. RL algorithms can be value-based (i.e., learning from off-policy experiences), policy-based (i.e., directly optimizing actions), or both. Value-based RL is guided by a $Q$-value function that indicates the goodness of taking an action given a specific state, guiding the agent in taking actions that maximize the $Q$-value \cite{Mnih2015}. In policy-based RL, however, the policy is updated directly, based on the gradient calculated from the reward parameters \cite{Sutton2000}, which is useful for problems with continuous action space.

\begin{figure}[ht]
  \centering \includegraphics[width=0.5\linewidth]{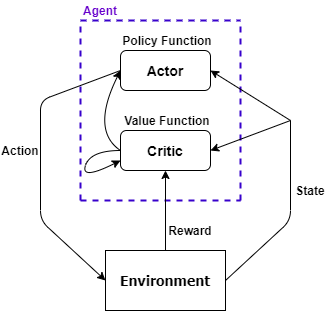}
    \caption{The actor-critic architecture.}
    \label{fig:AC}
\end{figure}

\paragraph{\textbf{Actor-critic architecture}} The actor-critic architecture builds on both value-based RL and policy-based RL, where a value function is used to update the policy parameters (Figure \ref{fig:AC}). A major drawback of off-policy learning is instability, which is addressed by the idea of \textit{target networks} to obtain more robust performance \cite{mnih2013playing}. In off-policy learning, the best performance is achieved when $\pi \approx \mu$, where $\mu$ is the behavior policy and $\pi$ is the target policy \cite{kapturowski2018recurrent}.
In the actor-critic framework \cite{degris2012off}, a $Q$-function acting as a critic for both the controller (i.e., ``actor'') and the value function (known as ``critic'') are learned simultaneously, as depicted in Figure \ref{fig:AC}.

In the actor-critic architecture, the policy is formulated as an approximate maximizer, i.e., $\pi_{\phi}(\mathbf{s}) \approx \arg \max _{\mathbf{a}} Q(\mathbf{s}, \mathbf{a})$. Therefore, the $Q$-function target (i.e., critic) is defined as $Q^{tar}=\sum_{i=0}^{n-1} \gamma^{i} r_{i}+\gamma^{n} Q_{\theta}(\mathbf{s}_{n}, \pi_{\phi}(\mathbf{s}_{n}))$. The critic and the policy (i.e., actor) are alternately updated with a learning rate $\alpha$, as follows:

\begin{equation}
    \theta^{\prime} \leftarrow \theta-\alpha \nabla_{\theta} \mathbb{E}_{\mathcal{D}}[(Q_{\theta}(\mathbf{s}_{0}, \mathbf{a}_{0})-Q^{tar})^{2}],
\end{equation}
\begin{equation}
    \phi^{\prime} \leftarrow \phi+\alpha \nabla_{\phi} \mathbb{E}_{\mathcal{D}}[Q_{\theta}(\mathbf{s}, \pi_{\phi}(\mathbf{s}))],
\end{equation}
where $\theta$ is soft $Q$-function network parameter, and $\phi$ is policy network parameter.

\paragraph{\textbf{Soft Actor-Critic (SAC) architecture}} It is a policy gradient algorithm based on combination of neural networks function approximation, double Q-learning \cite{van2016deep}, and entropy-regularized rewards to produce an off-policy actor-critic algorithm. The policy gradient objective of SAC (known as the maximum entropy RL objective) includes an entropy term $\mathbb{H}(\pi(\cdot| \mathbf{s}))$ as a regularizer \cite{ziebart2008maximum}:

\begin{equation}
J(\pi)= \sum_{t=0}^{T} \underset{\mathbf{s}_{t} \sim p, \mathbf{a}_{t} \sim \pi}{} \mathbb{E} \bigg[r(\mathbf{s}_{t}, \mathbf{a}_{t})+\alpha^{\prime} \mathbb{H}(\pi(\cdot | \mathbf{s})) \bigg],
\end{equation}
where the temperature parameter $\alpha^{\prime}$ is the ratio of the expected original task reward to the expected policy entropy, which balances exploitation and exploration.

SAC requires three functions to learn, which are associated with the actor and the critic. For critic, the action value function, $Q_{\theta}^{\pi}(\mathbf{s}, \mathbf{a})$, learns with minimizing a mean squared bootstrapped estimate (MSBE), and value function, $V_{\psi}^{\pi}(
\mathbf{s})$, learns with minimizing the squared residual error. In actor, target policy, $\pi$,
is a reparametrized Gaussian with the following squashing function:
\begin{equation}
\mathbf{a}_{t}=f_{\theta}(\mathbf{s}_{t} ; \epsilon_{t})=\tanh (\mu_{\theta}(\mathbf{s}_{t})+\sigma_{\theta}(\mathbf{s}_{t}) \cdot \epsilon_{t}),
\end{equation}
where $\epsilon_{t} \sim \mathcal{N}(0, I)$ is an input noise vector. Then, those parameters learn with maximizing the bootstrapped entropy regularized reward.

For training $\phi$ and $\theta$, a batch of transition tuples randomly samples from replay buffer $\mathcal{D}$, then stochastic gradient decent utilizes to minimize the following loss objectives:
\begin{equation}
L_{Q}(\theta):=\mathbb{E}_{\mathbf{s}_{t}, \mathbf{a}_{t} \sim \mathcal{D}}[\frac{1}{2} (Q_{\theta}(\mathbf{s}_{t}, \mathbf{a}_{t})-Q^{\operatorname{tar}}(\mathbf{s}_{t}, \mathbf{a}_{t}))^{2}], 
\end{equation}
where 
\begin{equation}
\begin{split}
&Q^{\operatorname{tar}}(\mathbf{s}_{t}, \mathbf{a}_{t}):=r(\mathbf{s}_{t}, \mathbf{a}_{t})+\gamma \mathbb{E}_{\mathbf{a}_{t+1} \sim p_{\phi}}[\hat{Q}(\mathbf{s}_{t+1}, \mathbf{a}_{t+1}) \\
&-\alpha \log (\pi_{\phi}(\mathbf{a}_{t+1} \mid \mathbf{s}_{t+1}))],
\end{split}
\end{equation}
\begin{equation}
L_{\pi}(\phi):=\mathbb{E}_{\mathbf{s}_{t} \sim \mathcal{D}, \mathbf{a}_{t} \sim \pi_{\phi}}[\alpha \log \pi_{\phi}(\mathbf{a}_{t} \mid \mathbf{s}_{t})-Q_{\theta}(\mathbf{s}_{t}, \mathbf{a}_{t})],
\end{equation}
where the parameters of $Q$-function are copied from $Q_{\theta}$ with target $\hat{Q}$.

Several algorithms have been developed to deal with the problem of high-dimensional continuous action spaces, as explained in Section \ref{background}. Further, several studies have been conducted on exploitation and exploration trade-offs; e.g., by employing entropy terms \cite{mnih2016asynchronous, haarnoja2018soft}. Among all existing continuous action RL algorithms, SAC is shown to achieve state-of-the-art performance on various robotics tasks (\cite{todorov2012mujoco}).

\section{Framework}
\label{framework}

The proposed Hyper-Actor Soft-Actor Critic (HASAC) framework is presented in this section. HASAC aims at improving the adaptability of the RL agent (e.g., a robot) to a wide range of tasks through task modularization, parameter-sharing, multi-task learning, and enhanced exploration-exploitation trade-off. The proposed framework builds on SAC as the state-of-the-art continuous control RL algorithm, and also utilizes SAC as a baseline for evaluating HASAC. SAC operates by updating three sets of parameters as follows: (1) the state value function network, $\psi$, (2) the soft $Q$-function network parameter, $\theta$, and (3) the policy network parameter, $\phi$. As a result, the tractable policy $\pi_{\phi}(\mathbf{a}_{t} | \mathbf{s}_{t})$ will also be updated. In the actor, the policy parameter, $\pi_{\phi}$, is learned directly by minimizing the expected Kullback-Leibler (KL) divergence:
\begin{equation}
J_{\pi}(\phi)=\mathbb{E}_{\mathbf{s}_{t} \sim \mathcal{D} \cup \mathcal{D}_{e}} \bigg[\operatorname{D_{KL}} (\pi_{\phi} (\cdot | \mathbf{s}_{t}) \| \frac{\exp (Q_{\theta}(\mathbf{s}_{t}, \cdot))}{Z_{\theta}(\mathbf{s}_{t})}) \bigg],
\end{equation}
where $\mathcal{D}$ is the distribution of previously sampled states and actions, called replay buffer. 

To efficiently balance the exploration and exploitation in the proposed HASAC, another replay buffer named \textit{elite reply buffer} ($\mathcal{D}_{e}$), is defined to collect states and actions that pass the reward function threshold ($\zeta$). The proposed mechanism leads to faster learning by improving the exploration and exploitation trade-off (for more detail, see the mechanism of replay buffers in \textbf{Algorithm \ref{HSAC}}).

To tackle complex tasks with continuous and high-dimensional action spaces, such as industrial collaborative robotics operations, HASAC aims at enabling the RL agent to automatically learn new tasks and transfer optimal policies across different tasks. These characteristics can be achieved by employing the aggregation concept defined based on the notions of modularity \cite{devin2017learning} and transfer learning \cite{barrett2010transfer}. This can be achieved either by sharing parameters across modules of a task or through multi-task learning. As the task becomes complex, current deep RL methods require to proportionally increase the training sample size to ensure desirable performance. However, question remains on how to train a single network that allows the sharing of some features across different robotic manipulation tasks, especially in real-world environments. 

Yu et al. \cite{yu2019meta} present a new robotic benchmark, Meta-World, to investigate the ideas of multi-task RL and meta RL on 50 different robotic manipulation tasks. They findings reveal that training agents on diverse robotic tasks affects the final performance compared to training them separately. We therefore speculate that defining a new general network for parameter sharing across tasks is expected to be a beneficial mechanism, allowing to update policy parameters while not affecting the optimization process of the task modules. This idea builds on the notion of transfer learning \cite{taylor2009transfer}, which aims at transferring the knowledge from learning an old task to a new task, to help the agent master new task faster. It is also applicable to learning at the task level when a task can be decomposed into several reusable modules \cite{devin2017learning}. 

\begin{figure}[t]
  \centering \includegraphics[width=1\linewidth]{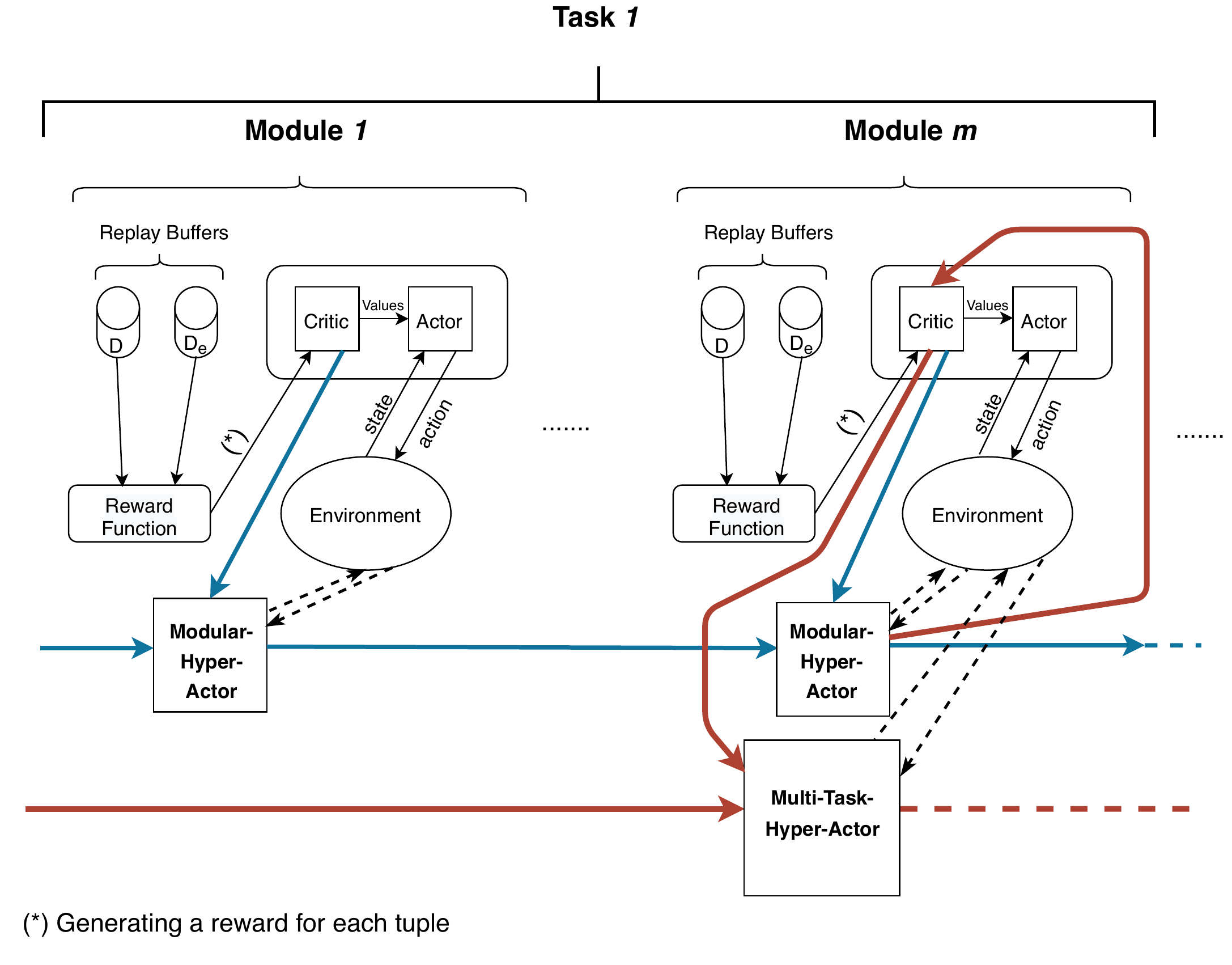}
    \caption{The proposed Hyper-Actor SAC (HASAC) framework: The hyper-actors transfer policies across different modules and tasks.}
    \label{fig:Diagram}
\end{figure}

The underlying idea of HASAC is based on the relationship between the actor and the critic networks (Figure \ref{fig:Diagram}). For each task, the critic updates the action-value function by computing the value function in an iterative fashion. Therefore, each task has a specific critic network. Since the output of the critic depends on the task, a separate critic is required for each task \cite{luo2016model}. In addition, each task module has its own critic, which helps increase the performance of the agent in learning the task faster. 

The actor also follows the standard actor-critic procedure, which involves learning the $Q$-value function for each state-action and updating the policy accordingly. Doing so, training a new single actor-network on all tasks, along with the current networks in the SAC, allows sharing the mutual features across different tasks. This is the idea which builds on the notion of transfer learning. Same as the critic network, the idea also can be applied on the module level. We refer to this new general actor as the ``hyper-actor''. In \textbf{Algorithm \ref{HSAC}}, \textit{$\mathcal{T}$} denotes the hyper-actor network that is trained on all interrelated tasks in multi-task learning and \textit{$\mathcal{M}_{i}$} denotes the hyper-actor network for all modules of the task $i$. As mentioned before, each module or task also has its specific actor-network from the baseline structure. Because, the trained actor on the baseline algorithm may receive different updates for the parameters. Therefore, two sets of parameters are introduced: $\mathbf{\varrho}^{\mathcal{M}_{i}}$ (\textit{Hyper-Actor-Module} parameters) and $\mathbf{\varrho}^{\mathcal{T}}$ (\textit{Hyper-Actor-Task} parameters). For more detail regarding the algorithm, see \textbf{Algorithm \ref{HSAC}}. See also the Nomenclature for notations. 

\begin{algorithm}[!ht]
\textbf{Input}: initial policy  parameter, $\phi$, soft $Q$-function network parameter, $\theta_1$, $\theta_2$, empty reply buffer $\mathcal{D}$, empty elite reply buffer $\mathcal{D}_{e}$,and $\pi_{\phi}(\mathbf{s})=\arg\max_a{Q(\mathbf{s},\mathbf{a})}$.\\
Set target parameters equal to main parameters $\psi_{\mathrm{targ}, 1} \leftarrow \psi_{1},  \psi_{\mathrm{targ}, 2} \leftarrow \psi_{2}$.\\
$\mathcal{T}$ is the set of tasks, $\mathcal{M}_{i}$ is the module set for task $i$.\\

Initialize Hyper-Actor-Task $\pi_{\phi}(\mathbf{s}|{\varrho^{\mu}}^{\mathcal{T}})$\\
\For{$i \in \mathcal{T}$}{
Initialize Hyper-Actor-Module $\pi_{\phi}(\mathbf{s}|{\varrho^{\mu}}^{\mathcal{M}_i})$\\
\For{$m \in \mathcal{M}_{i}$}{
\For{each iteration}{
\For{each environment step}{
$\mathbf{a}_{t} \sim \pi_{\phi}(\mathbf{a}_{t} | \mathbf{s}_{t})$ \\
$\mathbf{s}_{t+1} \sim p(\mathbf{s}_{t+1} | \mathbf{s}_{t}, \mathbf{a}_{t})$ \\
\eIf{$r(\mathbf{s}_{t}, \mathbf{a}_{t}) > \epsilon$}{
   $\mathcal{D}_{e} \leftarrow \mathcal{D}_{e} \cup\{(\mathbf{s}_{t}, \mathbf{a}_{t}, r(\mathbf{s}_{t}, \mathbf{a}_{t}), \mathbf{s}_{t+1})\}$
   }{
   $\mathcal{D} \leftarrow \mathcal{D} \cup\{(\mathbf{s}_{t}, \mathbf{a}_{t}, r(\mathbf{s}_{t}, \mathbf{a}_{t}), \mathbf{s}_{t+1})\}$
  }
}
\For{each gradient step}{
$\psi \leftarrow \psi-\lambda_{V} \hat{\nabla}_{\psi} J_{V}(\psi)$ \\
$\theta_{i} \leftarrow \theta_{i}-\lambda_{Q} \hat{\nabla}_{\theta_{i}} J_{Q}(\theta_{i}) \text { for } i \in\{1,2\} $\\
$\phi \leftarrow \phi-\lambda_{\pi} \hat{\nabla}_{\phi} J_{\pi}(\phi)$ \\
$\psi_{\mathrm{targ}, i} \leftarrow \tau \psi+(1-\tau) \psi_{\mathrm{targ}, i} \text { for } i \in\{1,2\}$
}
}
}
Save and free the Hyper-Actor-Module memory
}
\textbf{return}  Hyper-Actor-Task $\mu(\mathbf{s}|{\varrho^{\mu}}^{\mathcal{T}})$
\caption{HASAC}
\label{HSAC}
\end{algorithm}

Consider, a an example, the ``open window'' task of Meta-World \cite{yu2019meta} which contains ``reach'' and ``pull lever'' modules. At the module level, when the training phase for module ``reach'' finishes, the actor and critic parameters of the ``reach'' module will be reset. However, when the training process of the Hyper-Actor-Module network finishes, the memory will not be reset, and Hyper-Actor-Module parameters, $\mathbf{\varrho}^{\mathcal{M}_{reach}}$, will be transferred to the second module, ``pull lever''. Thus, the Hyper-Actor-Module is trained with the pull module's loss function, and then it will be trained on other modules. Since each module of a task has a specific loss function for the critic, it is an efficient way for only training a hyper-actor network in the test phase. Now suppose the ``open window'' task to be the first task in the task sequence of a multi-task learning problem. The actor of the ``open window'' is updated using the ``open window'' critic loss function, and then the value function of the ``open window'' critic will be used to update the Hyper-Actor-Module parameters ($\mathbf{\varrho}^{\mathcal{M}_{open-drawer}}$). Then, the updated $\mathbf{\varrho}^{\mathcal{M}_{open-drawer}}$ will be used for updating the loss function of the ``close window'' critic. Next, at the task level, the updated ``close window'' critic will be used to update the actor parameters of the Hyper-Actor-Task ($\mathbf{\varrho}^{\mathcal{T}}$) (see Figure \ref{fig:Diagram}).

The main difference between the Hyper-Actor-Module and Hyper-Actor-Task networks and the other networks is in the backpropagation process. The Hyper-Actor-Module and Hyper-Actor-Task networks are backpropagated \textit{cumulatively}. In multi-task learning, this implies that those networks are not freed within their training loop. Therefore, backpropagation will be continued for the multi-task network until the end of training tasks in the sequence of multi-task learning, which is the final \textit{Meta-Action} design for our testing experiment. Likewise, at the task level, the hyper-network is not freed within a task loop, and it will only be freed and saved once all modules of a task are trained. Hence, the hyper-network with parameter set $\mathbf{\varrho}^{\mathcal{M}_{i}}$ is the final action design for a particular task $i$. 

The backpropagation process of HASAC is similar to the process used for training the actor and critic networks in the SAC algorithm. Hence, two sets of networks, $\mathcal{M}_{i}$ and $\mathcal{T}$, are trained to enable a comprehensive learning framework which can be used for generalizing to new tasks at test time (see \textbf{Algorithm \ref{HSAC}}). Further, for the tasks that can be divided into different modules, the actor and critic networks parameters will be initialized with the policy parameters of the most related module from the past ($\phi \gets \varrho^{\mathcal{M}_{i}}$).By applying the proposed idea based on the relationship between the actor and critic networks as well as the notions of task modularization and transfer learning, HASAC can achieve significant improvements in both sample efficiency and final performance over previous state-of-the-art approaches, as demonstrated next.


\section{Experiments}
\label{experiments}

\begin{figure}[ht]
  \centering \includegraphics[width=0.8\linewidth]{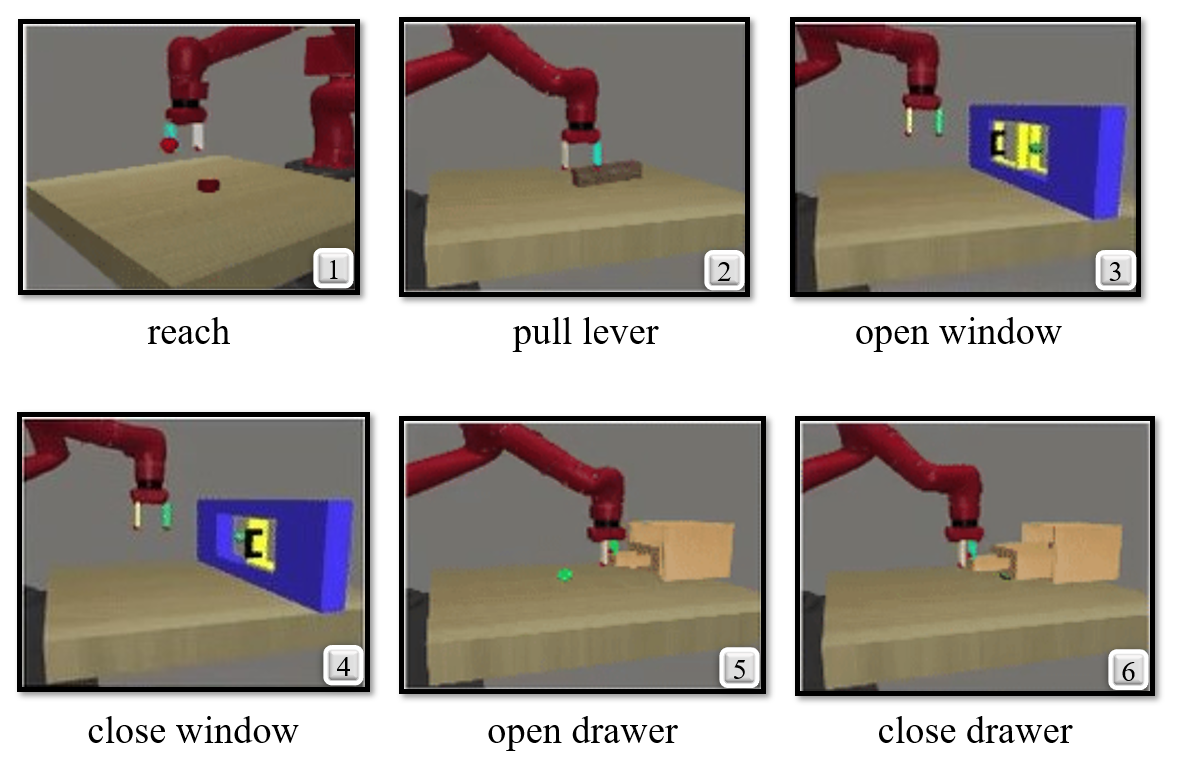}
    \caption{An example of visualization of the multi-task learning protocols. It demonstrates the training sequence of the modules and tasks for sharing parameters.}
    \label{fig:Protocol}
\end{figure}

This section evaluates the performance of the proposed HASAC framework on a set of robotic manipulation tasks from the Meta-World benchmark \cite{yu2019meta}. The Meta-World benchmark contains 50 different robot arm control tasks enabled by the MuJoCo physics simulator \cite{todorov2012mujoco}. All of the Meta-World tasks involve moving a simulated Sawyer robot arm to interact with objects by moving them from an initial state to a goal state. The action space consists of four degrees of freedom (three translational, one rotational) for the end-effector of the robot arm with observation space of 3D Cartesian positions of the object, the robot end-effector, and the goal position \cite{yun2020evaluating}. One of the essential steps in designing architecture is to properly modularize the tasks. Without loss of generality, our experiments are focused on four tasks ``open window'', ``close window'', ``open drawer'', and ``close drawer''. Although these tasks are symbolic and relatively simple, we believe analysis using the HASAC framework can be informative for solving realistic and more complex industrial problems. To modularize the aforementioned tasks, ``reach'' and ``pull lever'' are selected as the respective task modules (see Figure \ref{fig:Protocol}).

\subsection{Design of Experiments}
The open-source robotics environments of Meta-World enable us to study and analyze a set of virtual goal-oriented tasks via Mujoco Physics simulator \cite{todorov2012mujoco}. We compare the performance of HASAC with the standard SAC as a baseline. All algorithms were implemented using PyTorch. The experiments were performed on a computer with CPU AMD Ryzen Threadripper 2970WX 24-Core Processor 3.00GHZ. Table \ref{tab:parameters} demonstrates the hyperparameters used for running the experiments.

\begin{table}[ht]
\centering
\small
\caption{Hyperparameters setting.}
\begin{tabular}{ll}
\hline
      \textbf{Hyperparameters} & \textbf{Values}\\
\hline
      Optimizer & Adam\\ \hline
      Actor and critic networks: & 3 layers with 256 units\\
      & each and ReLU non-linearities   \\ \hline
      Learning rate of the actor & 0.001\\ \hline
      Learning rate of the critic & 0.001\\ \hline
      Discount factor ($\gamma$) & 0.98 \\ \hline
      Buffer size & $10^{6}$ transitions\\ \hline
      Action L2 norm coefficient &  1.0 \\ \hline
      Observation clipping & [-200, 200] \\ \hline
      Batch size & $10^{4}$ \\ \hline
      Epochs to train the agent &  100 \\ \hline
      Cycles per epoch &  50 \\ \hline 
      Batches per cycle & 40 \\ \hline
      Probability of random actions &  0.3 \\ \hline
      Test rollouts per epoch &  10 \\ \hline
      Scale of additive Gaussian noise & 0.2 \\ \hline
      Normalized clipping &  [-5, 5] \\ 
      \hline
\end{tabular}
\label{tab:parameters}
\end{table}

To successfully evaluate the performance of the proposed HASAC framework, proper metrics are required to test how successful each policy is. One of the metrics used in our experiments is the \textit{average reward} for the tasks. \textit{Success rate} in executing the tasks is another metric used in the experiments, defined based on the distance between the task-relevant object and its final goal position. Further, \textit{jumpstart} and \textit{asymptotic performance} are used as metrics to illustrate the benefits of transfer, as suggested by \cite{taylor2009transfer}. For training using the modularized HASAC algorithm, the first step is to pre-train the networks on a set of modules and tasks, and then use the pre-trained networks to initialize similar modules of a new task. Consequently, the agent is allowed to interact with the new environment and the network parameters are updated accordingly.

\subsection{Results and Analyses}
\begin{figure}[ht]
  \centering \includegraphics[width=1\linewidth]{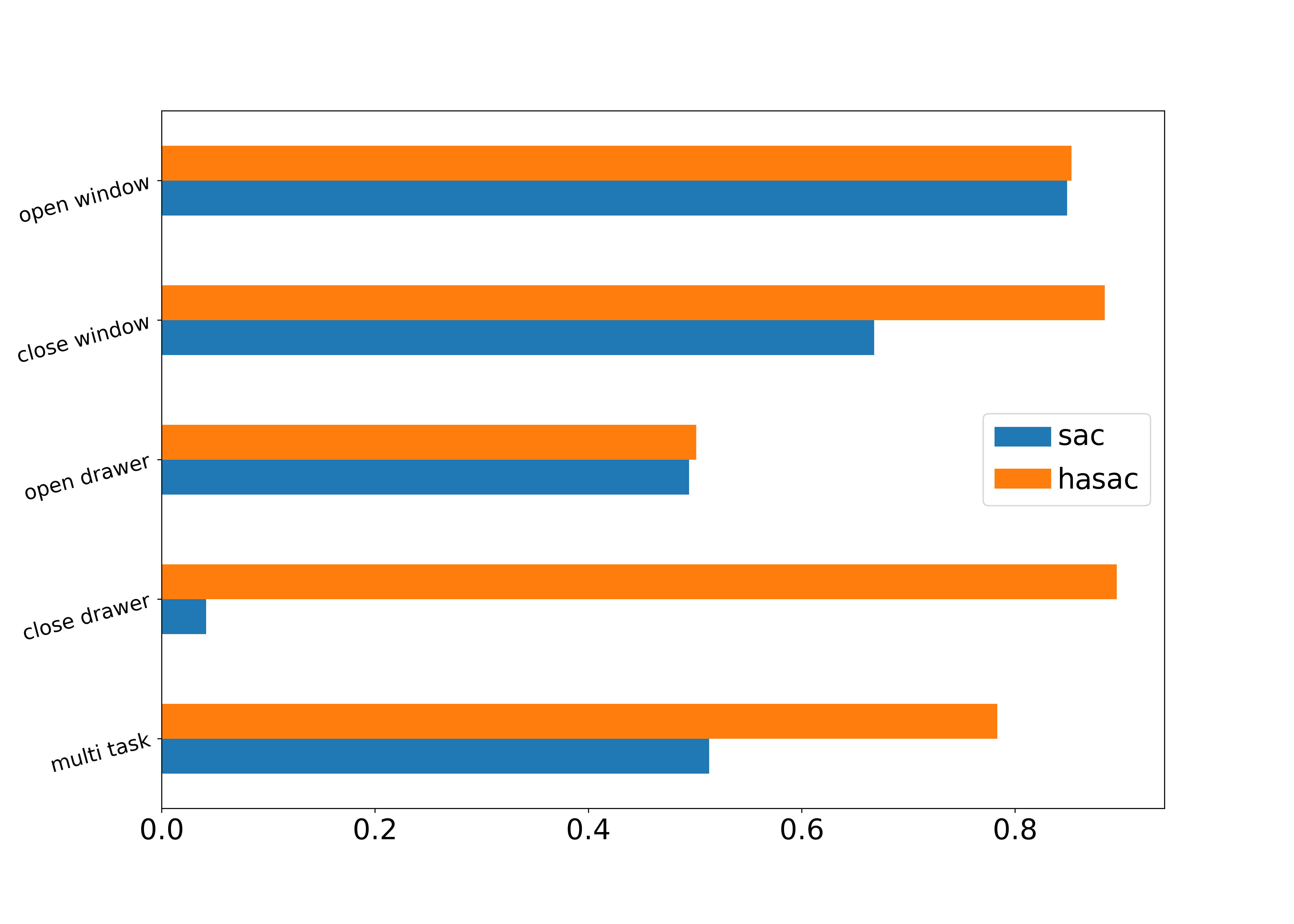}
    \caption{Comparison of average success rates for the selected tasks and multi-task learning.}
    \label{fig:successbar}
\end{figure}

To evaluate the performance of the HASAC framework on the four selected Meta-World tasks (i.e., ``open window'', ``close window'', ``open drawer'', and ``close drawer''), the tasks were first decomposed into modules, in the sequence of multi-task learning, and treated as sub-tasks associated with before and after reaching the object. First, the ``reach'' and ``pull lever'' modules were trained for each task on the HASAC, and the trained networks were then used for training each manipulation task. The tasks were also trained through multi-task learning. Figure \ref{fig:successbar} compares the average success rate for training four modularized tasks and multi-task learning. Using the modularized network and through transferring parameters across tasks in multi-task learning, the success rate is shown to improve compared to the standard SAC. Therefore, the results show that the proposed architecture leads to significant improvement in training modularized tasks and multi-task learning.

\begin{figure}[!ht]
\begin{minipage}[t]{0.49\linewidth}
    \centering
    \includegraphics[width=1\textwidth]{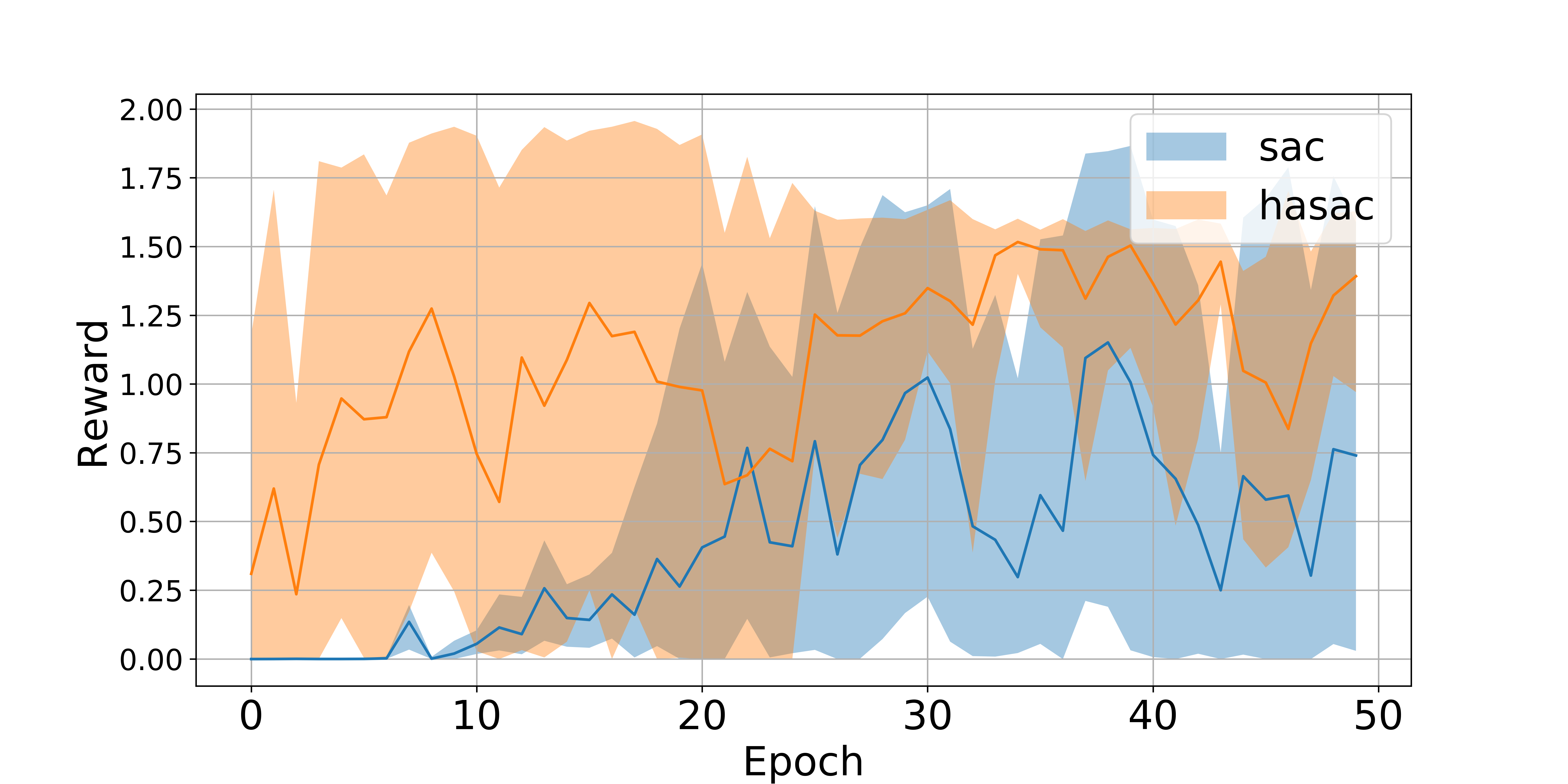}
    \caption{Comparison of reward function values for ``open window''.}
    \label{RetOpWin}
\end{minipage}
\hspace{0.1cm}
\begin{minipage}[t]{0.49\linewidth} 
    \centering
    \includegraphics[width=1\textwidth]{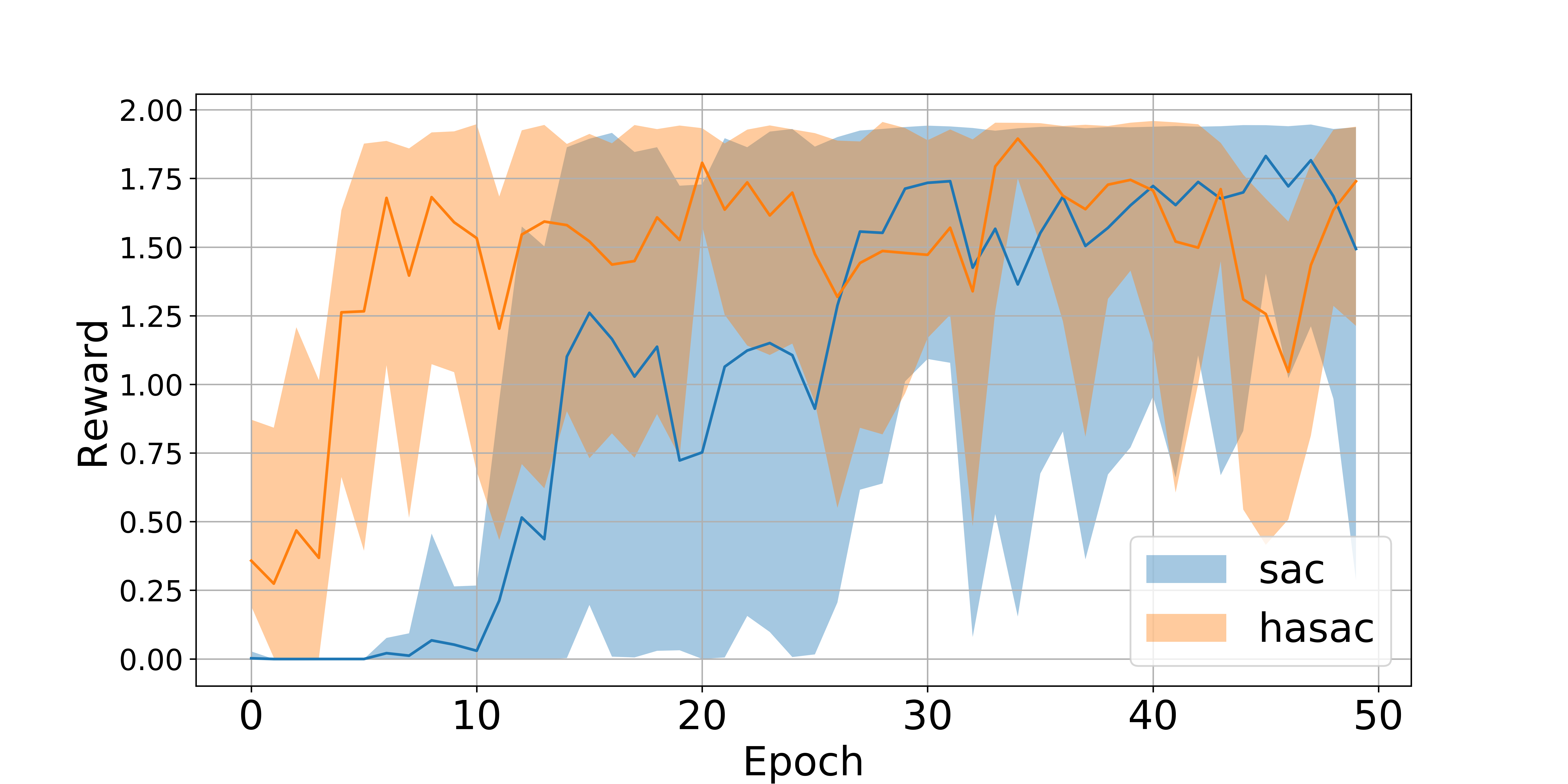}
    \caption{Comparison of reward function values for ``close window''.}
    \label{RetCloWin}
\end{minipage}        
\end{figure}

\begin{figure}[!ht]
\begin{minipage}[t]{0.49\linewidth}
    \centering
    \includegraphics[width=1\textwidth]{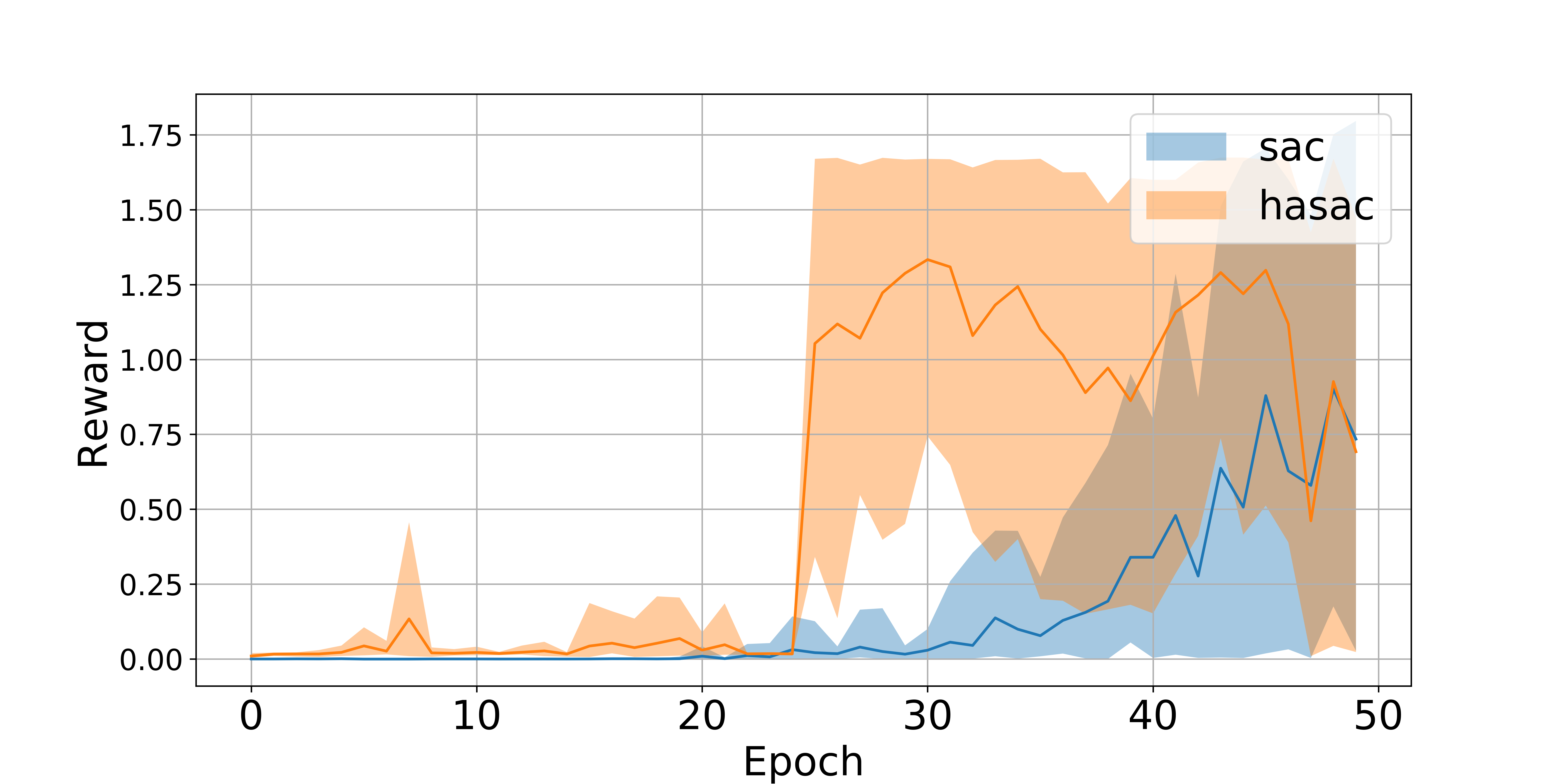}
    \caption{Comparison of reward function values for ``open drawer''.}
    \label{RetOpDra}
\end{minipage}
\hspace{0.1cm}
\begin{minipage}[t]{0.49\linewidth} 
    \centering
    \includegraphics[width=1\textwidth]{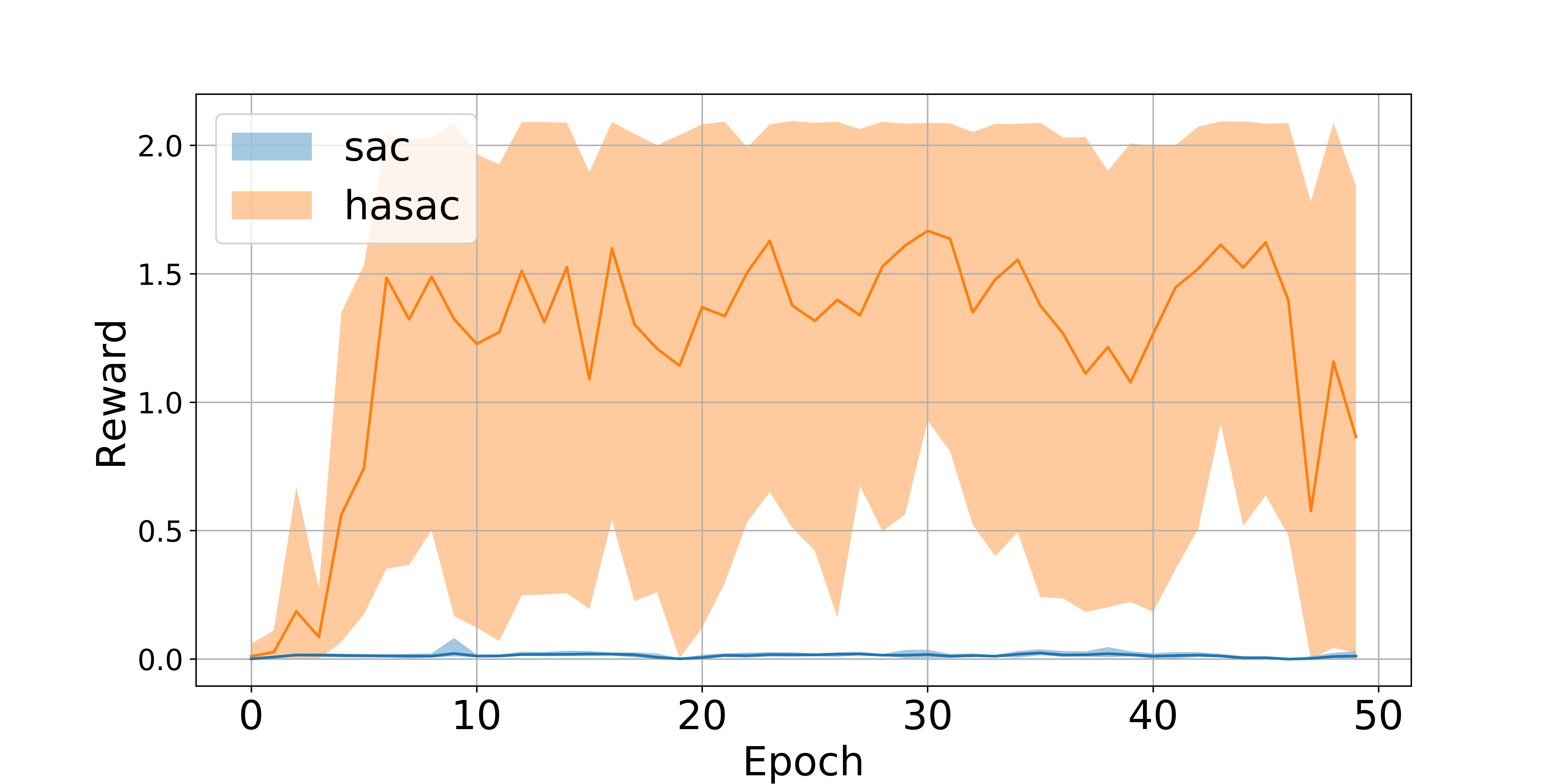}
    \caption{Comparison of reward function values for ``close drawer''}
    \label{RetCloDra1}
\end{minipage}        
\end{figure}

The reward function values depicted in Figures \ref{RetOpWin}-\ref{RetCloDra1} show that HASAC leads to significantly higher average reward than the baseline SAC. This is more evident for the ``close drawer'' task, which is the last task in the sequence of tasks trained through multi-task learning. In the ``close window'' and ``open window'' tasks, although both HASAC and SAC reach about the same reward function value, HASAC reaches its highest reward value relatively faster. Similar to the reward function value, the success rate of ``close drawer'' is significantly higher with the HASAC than with SAC (see Figure \ref{fig:successbar}). Moreover, it was observed that the success rate has a relatively lower variance under HASAC. This behavior results from introducing an inductive bias associated with transferring from previously learned task modules \cite{botvinick2019reinforcement}.

\begin{figure}[!ht]
\begin{minipage}[t]{0.49\linewidth}
    \centering
    \includegraphics[width=1\textwidth]{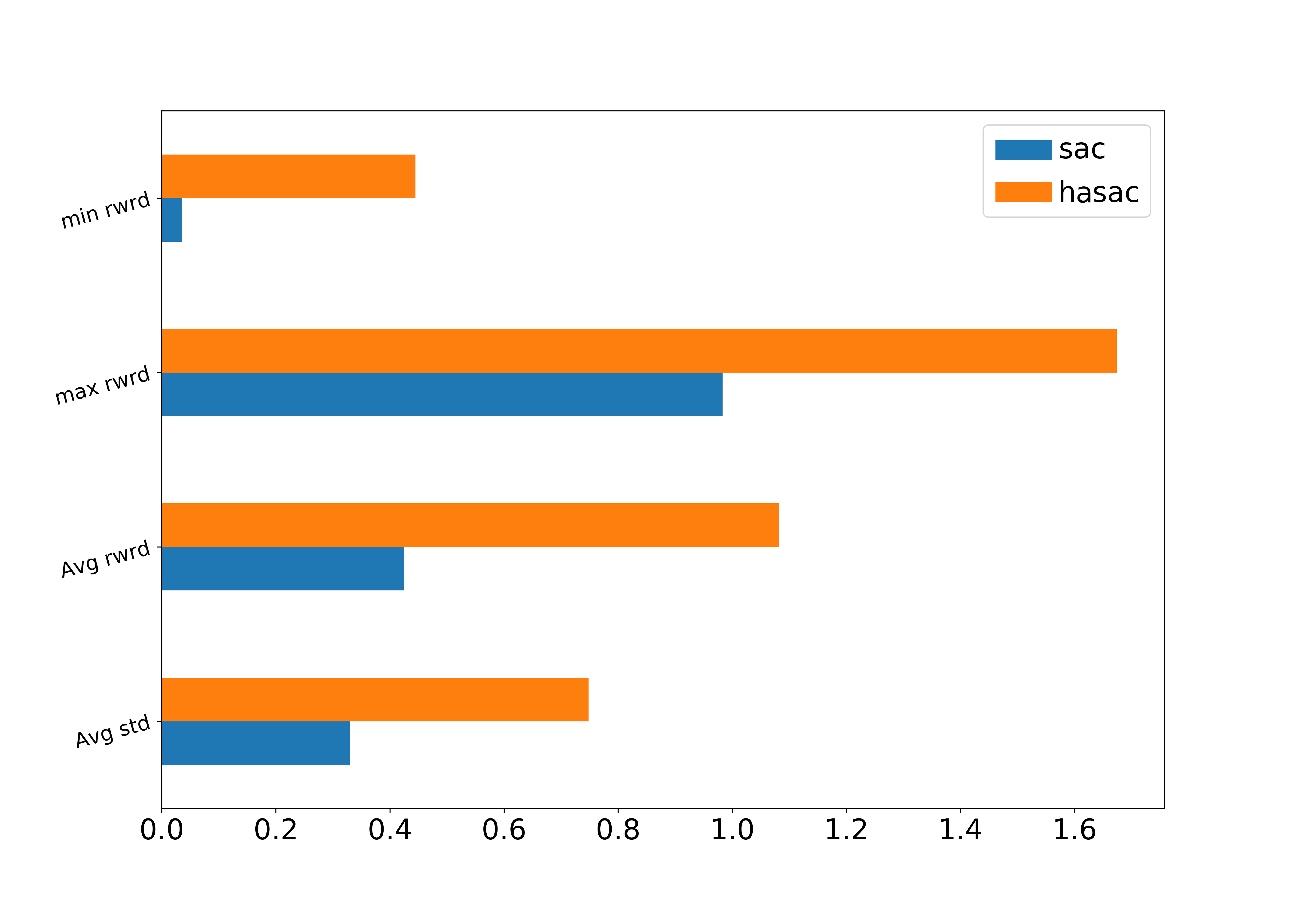}
    \caption{Overall comparison of reward function for ``open window''.}
    \label{BarOpWin}
\end{minipage}
\hspace{0.1cm}
\begin{minipage}[t]{0.49\linewidth} 
    \centering
    \includegraphics[width=1\textwidth]{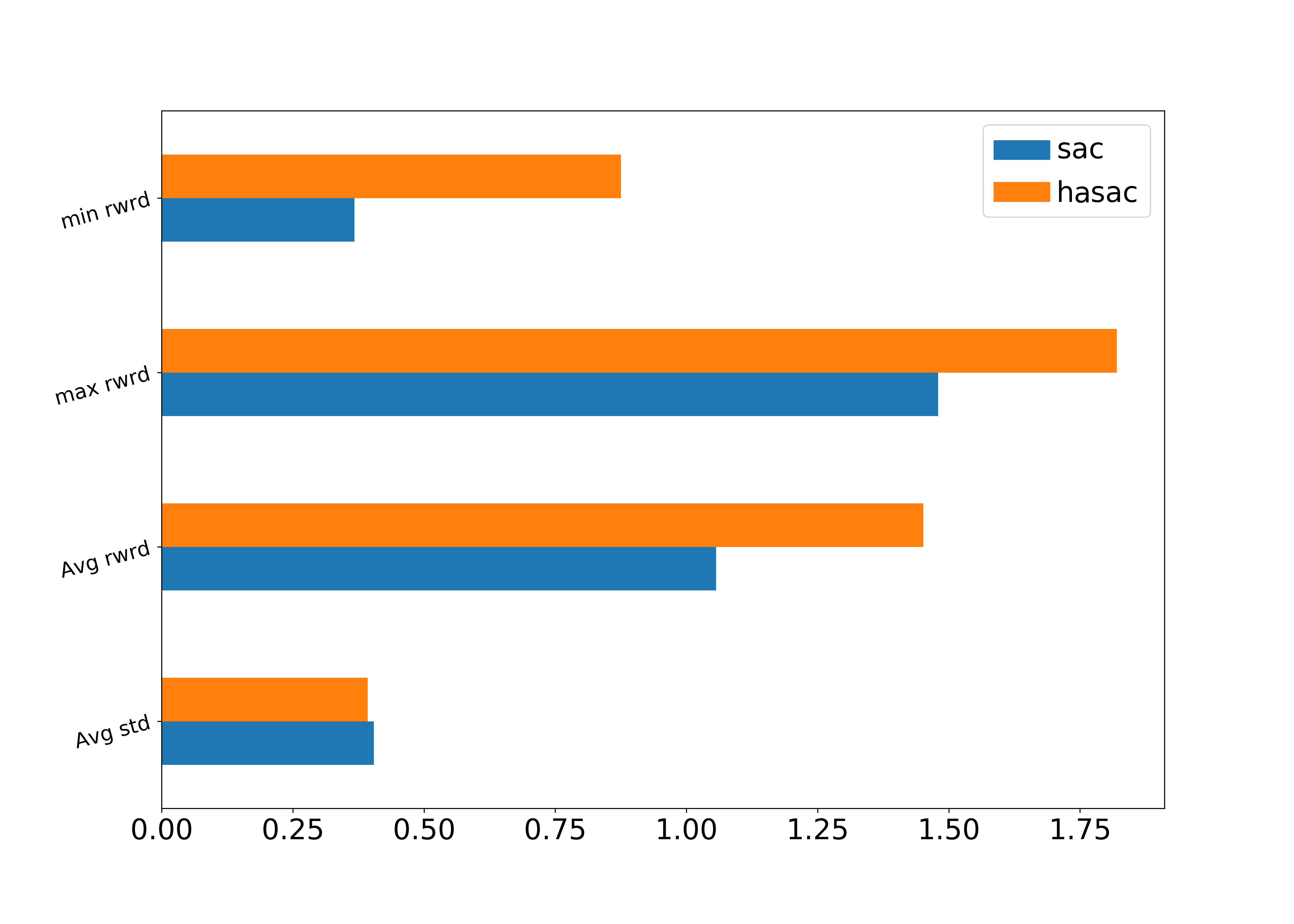}
    \caption{Overall comparison of reward function for ``close window''.}
    \label{BarCloWin}
\end{minipage}        
\end{figure}

\begin{figure}[!ht]
\begin{minipage}[t]{0.49\linewidth}
    \centering
    \includegraphics[width=1\textwidth]{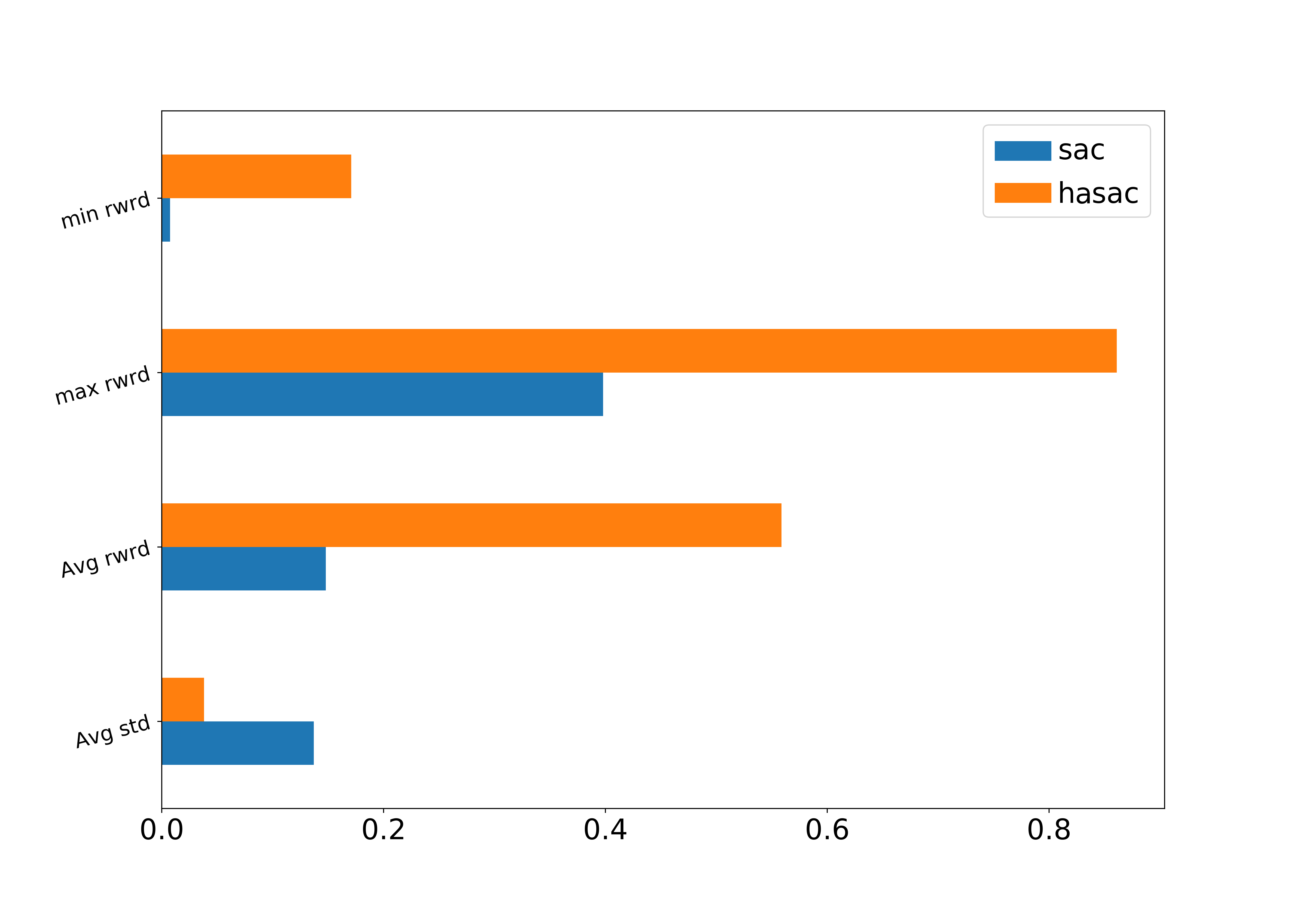}
    \caption{Overall comparison of reward function for ``open drawer''.}
    \label{BarOpDra}
\end{minipage}
\hspace{0.1cm}
\begin{minipage}[t]{0.49\linewidth} 
    \centering
    \includegraphics[width=1\textwidth]{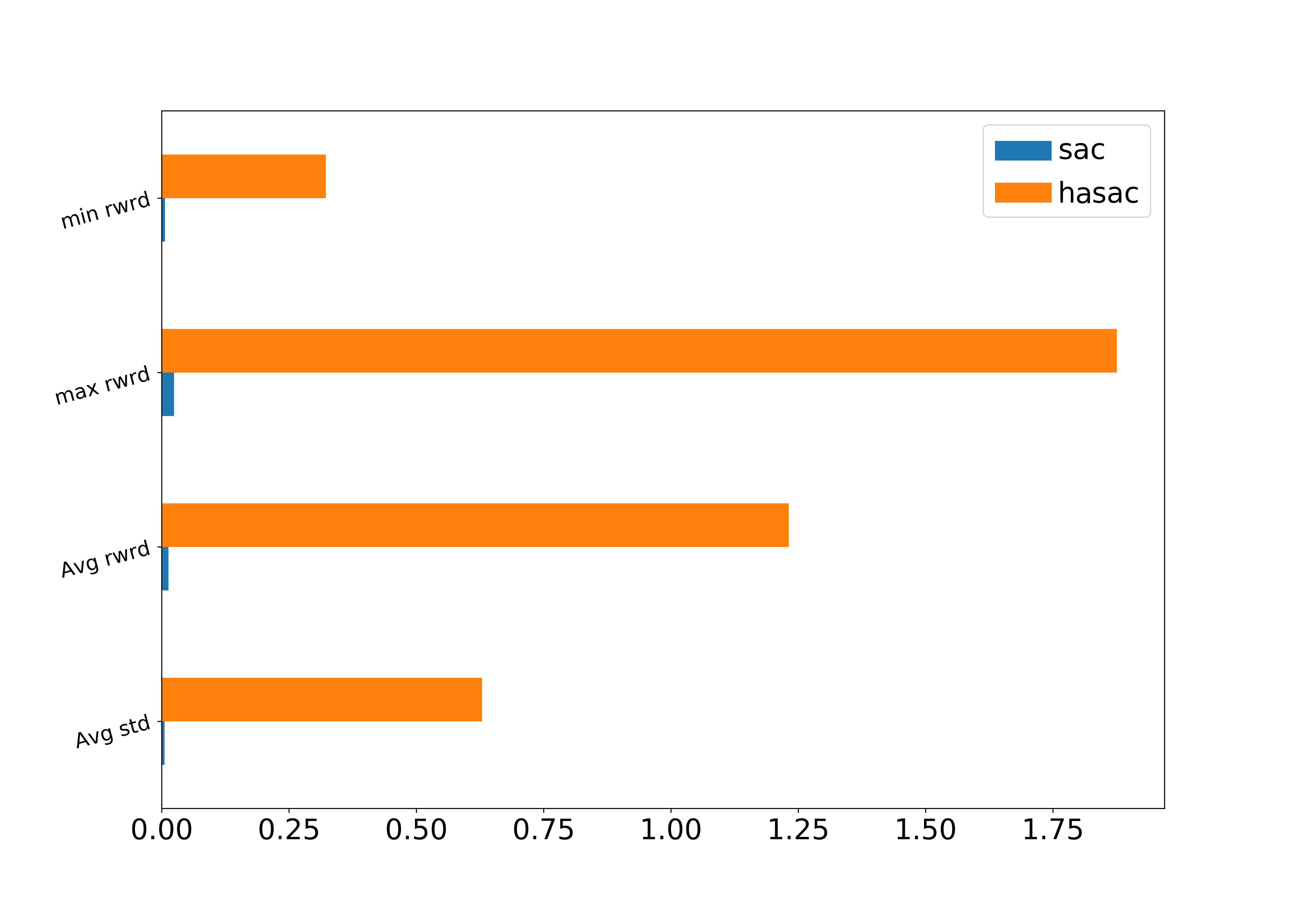}
    \caption{Overall comparison of reward function for ``close drawer''.}
    \label{RetCloDra}
\end{minipage}        
\end{figure}

The analysis of the reward function, presented in Figures \ref{BarOpWin}-\ref{RetCloDra} and Table \ref{tab:AverageReward}, illustrates that the HASAC enables a better trade-off between exploration and exploitation than the standard SAC. It is calculated in terms of the mean and standard deviation of the average reward function value for the selected tasks. Numerically, the results of HASAC show that the balance between two terms of the loss function leads to a better performance of the training algorithm.

\begin{table}
\centering
\small
\caption{Comparison of the average reward function values over 500 epochs.}
\begin{tabular}{|l|l|l|l|l|}
\hline
     $\times 10^6$     & Mean & STD & Max Reward & Min Reward      \\ \hline
          \hline
SAC & 0.4105 & 0.2192 & 0.7211 & 0.1041 \\ \hline
HSAC  & 1.1048 & 0.4519 & 1.5587 & 0.5028 \\ \hline
\end{tabular}
\label{tab:AverageReward}
\end{table}
\begin{figure}[ht]
  \centering \includegraphics[width=1\linewidth]{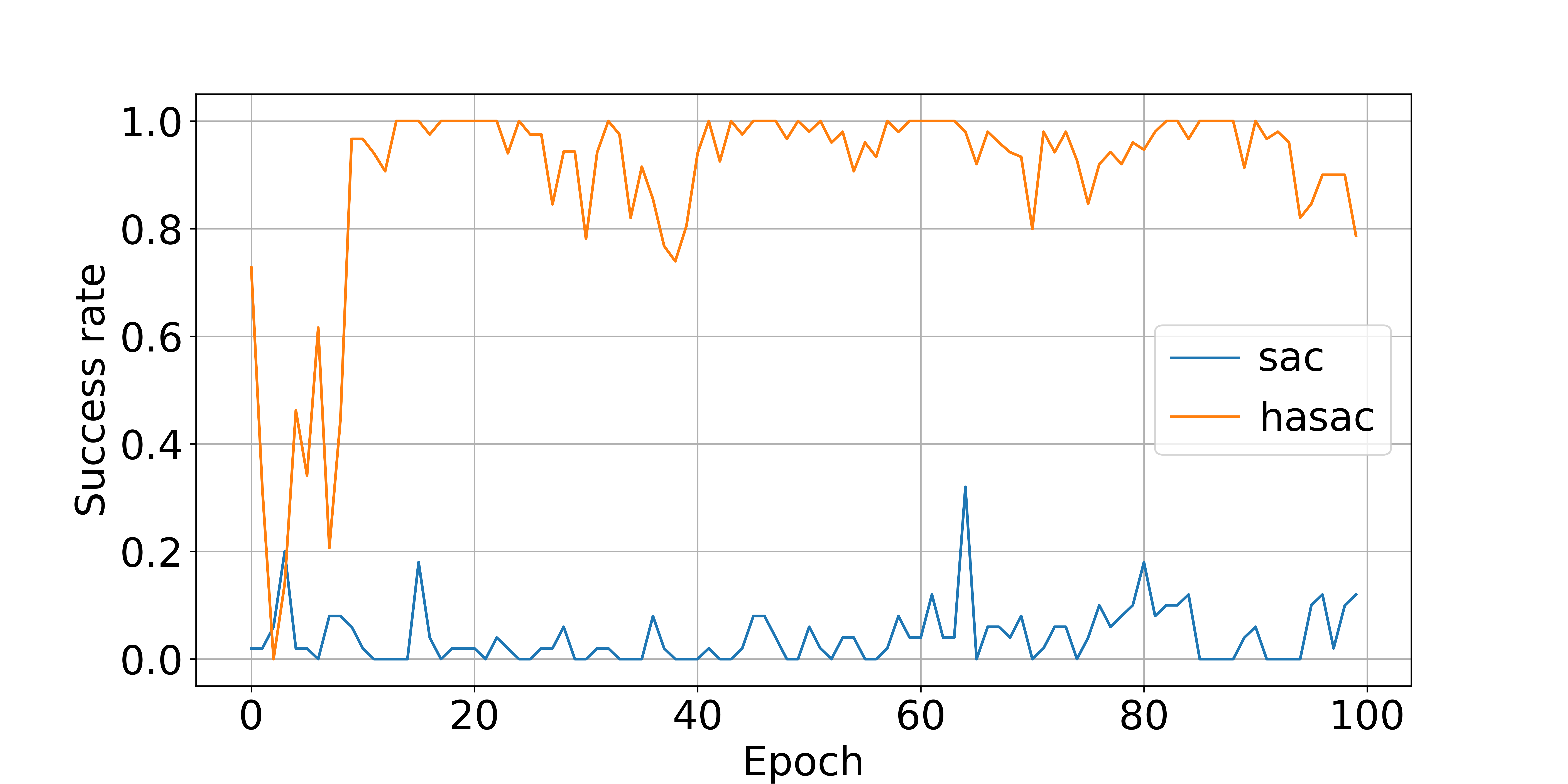}
    \caption{Comparison of success rate for final task in the sequence of multi-task learning.}
    \label{fig:succesCloDra}
\end{figure}
As discussed earlier, our proposed idea is based on the notions of task modularity and transfer learning. The experimental results presented and discussed in this section show that that the effectiveness of task modularity and transfer learning increases with increasing task complexity; which was the central hypothesis behind the HASAC framework. The results for metrics show that the simpler module affected by the hyper-actor parameters (in this case, ``pull lever'') receives more benefit associated with the initial jumpstart. In the long run, all trained modules and final trained tasks benefit from modularization (see Figure \ref{fig:succesCloDra}, the comparison result of success rate for final task in our experiments). Due to limited compute resources, however, we trained the network for 500 epochs, which is at least 50 times less than similar studies in the literature. The completion time indicates that the average length of the training time decreases as the robot's success rate improves. Both proposed metrics converge to an optimal solution. As a result, the proposed framework is expected to significantly improve multi-task and meta-learning algorithms on various robotic manipulation problems.

\section{Conclusions and Future Research Directions}
\label{conclusions}
This article addresses one of the key research questions associated with the adaptability of industrial automation systems to the increasing variability of tasks assigned to automated equipment such as collaborative robots. A novel modularized actor-critic based deep RL framework is developed and tested on multiple simulated robotic manipulation tasks. The main goal is to improve performance and sample efficiency for training robotic manipulation tasks and multi-task learning through task modularization and transfer of policy parameters using ``hyper-actor'' networks at both module level and task level. Experiments indicated that incorporating task modularization in actor-critic based RL methods is an effective mechanism to learn complex tasks faster and more efficiently. Besides, the experimental results concurred that the effectiveness of this framework increases with increasing task complexity, and also it enables the agent to learn simple tasks faster.

The experimental results also show that the proposed HASAC framework achieves better success rate and converges faster with more stability than the baseline SAC algorithm by a relatively large margin. Moreover, results show that the trained \textit{Hyper-Actor-Module} and \textit{Hyper-Actor-Task} networks improve the success rate of the final task across the sequence of multi-task learning. Indeed, the hyper-network architecture allows efficient sharing and reuse of network elements across tasks, which creates future opportunities for developing more robust and efficient meta RL algorithms. 

A major limitation of this work is the lack of experimentation on more complex and realistic tasks, both in simulated environments and on a real robot in a laboratory setting. This should be one of the foremost future research directions in this area. Future research should also investigate more meaningful industrial use-cases where the use of RL and AI in general is preferred to traditional approaches for programming automation equipment. One important example from industrial, collaborative robotic applications is human-robot collaboration where the presence of the human in the loop both justifies the incorporation of AI in robotic control and facilitates the task modularization and policy transfer processes. Future research should also investigate the implications of the proposed framework for a variety of industrial sectors including manufacturing, healthcare, defense, and agriculture, among others.

\section*{Nomenclature}

\begin{longtable}[l]{p{2cm} p{14cm}}
    $\mathbf{s} \in \mathcal{S}$ & \quad States \\
    $\mathbf{a} \in \mathcal{A}$ & \quad Actions \\
    $r \in \mathcal{R}$ & \quad Rewards \\
    $\mathcal{L}$ & \quad Loss function \\
    $\gamma$ & \quad Discount factor ($0<\gamma \leq 1$) \\
    $\alpha$ & \quad Learning rate \\
    $\pi$ & \quad Policy, $\pi_{\theta}(.)$, is a policy parameterized by $\theta$ \\
    $K$ & \quad Length of episode \\
    $\psi$ & \quad State value function network parameter\\
    $\theta$ & \quad $Q$-function network parameter\\
    $\phi$ & \quad Policy network parameter\\
    $V(\mathbf{s})$ & \quad State-value function which determines expected return of state\\
    $Q(\mathbf{s}, \mathbf{a})$ & \quad Action-value function which determines the expected return of\\
    & \quad a pair of state and action\\
    $ V_{\psi}(.)$ & \quad Value function parameterized by $\psi$\\
    $Q_{\theta}(.)$& \quad Action value function parameterized by $\theta$\\
    $V^{\pi}(\mathbf{s})$ & \quad Value of a state $\mathbf{s}$ when follows a policy $\pi$\\
    $Q^{\pi}(\mathbf{s}, \mathbf{a})$ & \quad Value of pair state $\mathbf{s}$ and action $\mathbf{a}$ when follows a policy $\pi$\\
    $p(\mathbf{s}_{t+1} | \mathbf{s}_{t}, \mathbf{a}_{t})$ & \quad Transition distribution over episode length $K$\\
    $p(\mathcal{T})$ & \quad Distribution over tasks, \\ 
    $\alpha^{\prime}$ & \quad Temperature parameter\\
    $\mathcal{D}$ & \quad Replay buffer\\
    $\mathcal{D}_{e}$ & \quad Elite reply buffer \\
    $\mathcal{M}_{i}$ & \quad Module set for task $i$\\
    $\mathcal{T}$ & \quad Set of tasks\\
    $\mathbf{\varrho}^{\mathcal{M}_{i}}$ & \quad Hyper-Actor-Module parameters \\
    $\mathbf{\varrho}^{\mathcal{T}}$ & \quad Hyper-Actor-Task parameter
\end{longtable}

\bibliography{mybibfile}

\begin{thebibliography}{10}
\expandafter\ifx\csname url\endcsname\relax
  \def\url#1{\texttt{#1}}\fi
\expandafter\ifx\csname urlprefix\endcsname\relax\def\urlprefix{URL }\fi
\expandafter\ifx\csname href\endcsname\relax
  \def\href#1#2{#2} \def\path#1{#1}\fi

\bibitem{koren1999reconfigurable}
Y.~Koren, U.~Heisel, F.~Jovane, T.~Moriwaki, G.~Pritschow, G.~Ulsoy,
  H.~Van~Brussel, Reconfigurable manufacturing systems, Annals of the CIRP 48
  (1999) 2.

\bibitem{trentesaux2009distributed}
D.~Trentesaux, Distributed control of production systems, Engineering
  Applications of Artificial Intelligence 22~(7) (2009) 971--978.

\bibitem{leitao2012bio}
P.~Leit{\~a}o, J.~Barbosa, D.~Trentesaux, Bio-inspired multi-agent systems for
  reconfigurable manufacturing systems, Engineering Applications of Artificial
  Intelligence 25~(5) (2012) 934--944.

\bibitem{moghaddam2018reference}
M.~Moghaddam, M.~N. Cadavid, C.~R. Kenley, A.~V. Deshmukh, Reference
  architectures for smart manufacturing: a critical review, Journal of
  manufacturing systems 49 (2018) 215--225.

\bibitem{yu2019reinforcement}
C.~Yu, J.~Liu, S.~Nemati, Reinforcement learning in healthcare: A survey, arXiv
  preprint arXiv:1908.08796 (2019).

\bibitem{ruiz2020predictive}
J.-R. Ruiz-Sarmiento, J.~Monroy, F.-A. Moreno, C.~Galindo, J.-M. Bonelo,
  J.~Gonzalez-Jimenez, A predictive model for the maintenance of industrial
  machinery in the context of industry 4.0, Engineering Applications of
  Artificial Intelligence 87 (2020) 103289.

\bibitem{weyer2015towards}
S.~Weyer, M.~Schmitt, M.~Ohmer, D.~Gorecky, Towards industry
  4.0-standardization as the crucial challenge for highly modular, multi-vendor
  production systems, Ifac-Papersonline 48~(3) (2015) 579--584.

\bibitem{arinez2020artificial}
J.~F. Arinez, Q.~Chang, R.~X. Gao, C.~Xu, J.~Zhang, Artificial intelligence in
  advanced manufacturing: Current status and future outlook, Journal of
  Manufacturing Science and Engineering 142~(11) (2020).

\bibitem{bhattacharya2016review}
S.~Bhattacharya, A review of the application of automation technologies in
  healthcare domain, Research Journal of Pharmacy and Technology 9~(12) (2016)
  2343--2348.

\bibitem{robert2017growing}
L.~Robert, The growing problem of humanizing robots, Robert, LP (2017). The
  Growing Problem of Humanizing Robots, International Robotics \& Automation
  Journal 3~(1) (2017).

\bibitem{marinoudi2019robotics}
V.~Marinoudi, C.~G. S{\o}rensen, S.~Pearson, D.~Bochtis, Robotics and labour in
  agriculture. a context consideration, Biosystems Engineering 184 (2019)
  111--121.

\bibitem{ivanov2020digital}
D.~Ivanov, A.~Dolgui, A digital supply chain twin for managing the disruption
  risks and resilience in the era of industry 4.0, Production Planning \&
  Control (2020) 1--14.

\bibitem{chou2018fourth}
S.-Y. Chou, The fourth industrial revolution, Journal of International Affairs
  72~(1) (2018) 107--120.

\bibitem{pane2019reinforcement}
Y.~P. Pane, S.~P. Nageshrao, J.~Kober, R.~Babu{\v{s}}ka, Reinforcement learning
  based compensation methods for robot manipulators, Engineering Applications
  of Artificial Intelligence 78 (2019) 236--247.

\bibitem{stone2005reinforcement}
P.~Stone, R.~S. Sutton, G.~Kuhlmann, Reinforcement learning for robocup soccer
  keepaway, Adaptive Behavior 13~(3) (2005) 165--188.

\bibitem{riedmiller2007learning}
M.~Riedmiller, M.~Montemerlo, H.~Dahlkamp, Learning to drive a real car in 20
  minutes, in: 2007 Frontiers in the Convergence of Bioscience and Information
  Technologies, IEEE, 2007, pp. 645--650.

\bibitem{sutton2018reinforcement}
R.~S. Sutton, A.~G. Barto, Reinforcement learning: An introduction, MIT press,
  2018.

\bibitem{bansal2017emergent}
T.~Bansal, J.~Pachocki, S.~Sidor, I.~Sutskever, I.~Mordatch, Emergent
  complexity via multi-agent competition, arXiv preprint arXiv:1710.03748
  (2017).

\bibitem{heess2017emergence}
N.~Heess, D.~TB, S.~Sriram, J.~Lemmon, J.~Merel, G.~Wayne, Y.~Tassa, T.~Erez,
  Z.~Wang, S.~Eslami, et~al., Emergence of locomotion behaviours in rich
  environments, arXiv preprint arXiv:1707.02286 (2017).

\bibitem{levine2016end}
S.~Levine, C.~Finn, T.~Darrell, P.~Abbeel, End-to-end training of deep
  visuomotor policies, The Journal of Machine Learning Research 17~(1) (2016)
  1334--1373.

\bibitem{duan1611rl2}
Y.~Duan, J.~Schulman, X.~Chen, P.~L. Bartlett, I.~Sutskever, P.~Abbeel, Rl2:
  Fast reinforcement learning via slow reinforcement learning. 2016, arXiv
  preprint arXiv:1611.02779 (2016).

\bibitem{tamar2017learning}
A.~Tamar, G.~Thomas, T.~Zhang, S.~Levine, P.~Abbeel, Learning from the
  hindsight plan—episodic mpc improvement, in: 2017 IEEE International
  Conference on Robotics and Automation (ICRA), IEEE, 2017, pp. 336--343.

\bibitem{yu2019meta}
T.~Yu, D.~Quillen, Z.~He, R.~Julian, K.~Hausman, C.~Finn, S.~Levine,
  Meta-world: A benchmark and evaluation for multi-task and meta reinforcement
  learning, arXiv preprint arXiv:1910.10897 (2019).

\bibitem{wang2016learning}
J.~X. Wang, Z.~Kurth-Nelson, D.~Tirumala, H.~Soyer, J.~Z. Leibo, R.~Munos,
  C.~Blundell, D.~Kumaran, M.~Botvinick, Learning to reinforcement learn, arXiv
  preprint arXiv:1611.05763 (2016).

\bibitem{xu2018meta}
Z.~Xu, H.~P. van Hasselt, D.~Silver, Meta-gradient reinforcement learning, in:
  Advances in neural information processing systems, 2018, pp. 2396--2407.

\bibitem{battaglia2018relational}
P.~W. Battaglia, J.~B. Hamrick, V.~Bapst, A.~Sanchez-Gonzalez, V.~Zambaldi,
  M.~Malinowski, A.~Tacchetti, D.~Raposo, A.~Santoro, R.~Faulkner, et~al.,
  Relational inductive biases, deep learning, and graph networks, arXiv
  preprint arXiv:1806.01261 (2018).

\bibitem{botvinick2019reinforcement}
M.~Botvinick, S.~Ritter, J.~X. Wang, Z.~Kurth-Nelson, C.~Blundell, D.~Hassabis,
  Reinforcement learning, fast and slow, Trends in cognitive sciences (2019).

\bibitem{lebensold2019actor}
J.~Lebensold, W.~Hamilton, B.~Balle, D.~Precup, Actor critic with
  differentially private critic, arXiv preprint arXiv:1910.05876 (2019).

\bibitem{humplik2019meta}
J.~Humplik, A.~Galashov, L.~Hasenclever, P.~A. Ortega, Y.~W. Teh, N.~Heess,
  Meta reinforcement learning as task inference, arXiv preprint
  arXiv:1905.06424 (2019).

\bibitem{dulac2020empirical}
G.~Dulac-Arnold, N.~Levine, D.~J. Mankowitz, J.~Li, C.~Paduraru, S.~Gowal,
  T.~Hester, An empirical investigation of the challenges of real-world
  reinforcement learning, arXiv preprint arXiv:2003.11881 (2020).

\bibitem{mnih2015human}
V.~Mnih, K.~Kavukcuoglu, D.~Silver, A.~A. Rusu, J.~Veness, M.~G. Bellemare,
  A.~Graves, M.~Riedmiller, A.~K. Fidjeland, G.~Ostrovski, et~al., Human-level
  control through deep reinforcement learning, Nature 518~(7540) (2015) 529.

\bibitem{silver2017alphago}
D.~Silver, D.~Hassabis, Alphago zero: Starting from scratch, London, United
  Kingdom: Deepmind (2017).

\bibitem{moravvcik2017deepstack}
M.~Morav{\v{c}}{\'\i}k, M.~Schmid, N.~Burch, V.~Lis{\`y}, D.~Morrill, N.~Bard,
  T.~Davis, K.~Waugh, M.~Johanson, M.~Bowling, Deepstack: Expert-level
  artificial intelligence in heads-up no-limit poker, Science 356~(6337) (2017)
  508--513.

\bibitem{he2007reinforcement}
P.~He, S.~Jagannathan, Reinforcement learning neural-network-based controller
  for nonlinear discrete-time systems with input constraints, IEEE Transactions
  on Systems, Man, and Cybernetics, Part B (Cybernetics) 37~(2) (2007)
  425--436.

\bibitem{duan2016benchmarking}
Y.~Duan, X.~Chen, R.~Houthooft, J.~Schulman, P.~Abbeel, Benchmarking deep
  reinforcement learning for continuous control, in: International Conference
  on Machine Learning, 2016, pp. 1329--1338.

\bibitem{lillicrap2015continuous}
T.~P. Lillicrap, J.~J. Hunt, A.~Pritzel, N.~Heess, T.~Erez, Y.~Tassa,
  D.~Silver, D.~Wierstra, Continuous control with deep reinforcement learning,
  arXiv preprint arXiv:1509.02971 (2015).

\bibitem{kim2020motion}
M.~Kim, D.-K. Han, J.-H. Park, J.-S. Kim, Motion planning of robot manipulators
  for a smoother path using a twin delayed deep deterministic policy gradient
  with hindsight experience replay, Applied Sciences 10~(2) (2020) 575.

\bibitem{haarnoja2018soft}
T.~Haarnoja, A.~Zhou, P.~Abbeel, S.~Levine, Soft actor-critic: Off-policy
  maximum entropy deep reinforcement learning with a stochastic actor, arXiv
  preprint arXiv:1801.01290 (2018).

\bibitem{zambaldi2018deep}
V.~Zambaldi, D.~Raposo, A.~Santoro, V.~Bapst, Y.~Li, I.~Babuschkin, K.~Tuyls,
  D.~Reichert, T.~Lillicrap, E.~Lockhart, et~al., Deep reinforcement learning
  with relational inductive biases, International Conference on Learning
  Representations (2018).

\bibitem{gupta2018unsupervised}
A.~Gupta, B.~Eysenbach, C.~Finn, S.~Levine, Unsupervised meta-learning for
  reinforcement learning, arXiv preprint arXiv:1806.04640 (2018).

\bibitem{jabri2019unsupervised}
A.~Jabri, K.~Hsu, A.~Gupta, B.~Eysenbach, S.~Levine, C.~Finn, Unsupervised
  curricula for visual meta-reinforcement learning, in: Advances in Neural
  Information Processing Systems, 2019, pp. 10519--10530.

\bibitem{alet2018modular}
F.~Alet, T.~Lozano-P{\'e}rez, L.~P. Kaelbling, Modular meta-learning, arXiv
  preprint arXiv:1806.10166 (2018).

\bibitem{taylor2009transfer}
M.~E. Taylor, P.~Stone, Transfer learning for reinforcement learning domains: A
  survey, Journal of Machine Learning Research 10~(Jul) (2009) 1633--1685.

\bibitem{zhang2017survey}
Y.~Zhang, Q.~Yang, A survey on multi-task learning, arXiv preprint
  arXiv:1707.08114 (2017).

\bibitem{rahmatizadeh2018vision}
R.~Rahmatizadeh, P.~Abolghasemi, L.~B{\"o}l{\"o}ni, S.~Levine, Vision-based
  multi-task manipulation for inexpensive robots using end-to-end learning from
  demonstration, in: 2018 IEEE International Conference on Robotics and
  Automation (ICRA), IEEE, 2018, pp. 3758--3765.

\bibitem{fox2019multi}
R.~Fox, R.~Berenstein, I.~Stoica, K.~Goldberg, Multi-task hierarchical
  imitation learning for home automation, in: 2019 IEEE 15th International
  Conference on Automation Science and Engineering (CASE), IEEE, 2019, pp.
  1--8.

\bibitem{pinto2017learning}
L.~Pinto, A.~Gupta, Learning to push by grasping: Using multiple tasks for
  effective learning, in: 2017 IEEE International Conference on Robotics and
  Automation (ICRA), IEEE, 2017, pp. 2161--2168.

\bibitem{gueant2019deep}
O.~Gu{\'e}ant, I.~Manziuk, Deep reinforcement learning for market making in
  corporate bonds: beating the curse of dimensionality, Applied Mathematical
  Finance 26~(5) (2019) 387--452.

\bibitem{silver2016mastering}
D.~Silver, A.~Huang, C.~J. Maddison, A.~Guez, L.~Sifre, G.~Van Den~Driessche,
  J.~Schrittwieser, I.~Antonoglou, V.~Panneershelvam, M.~Lanctot, et~al.,
  Mastering the game of go with deep neural networks and tree search, nature
  529~(7587) (2016) 484.

\bibitem{vinyals2019alphastar}
O.~Vinyals, I.~Babuschkin, J.~Chung, M.~Mathieu, M.~Jaderberg, W.~M. Czarnecki,
  A.~Dudzik, A.~Huang, P.~Georgiev, R.~Powell, et~al., Alphastar: Mastering the
  real-time strategy game starcraft ii, DeepMind blog (2019) 2.

\bibitem{garcia2020teaching}
J.~Garc{\'\i}a, D.~Shafie, Teaching a humanoid robot to walk faster through
  safe reinforcement learning, Engineering Applications of Artificial
  Intelligence 88 (2020) 103360.

\bibitem{pan2009survey}
S.~J. Pan, Q.~Yang, A survey on transfer learning, IEEE Transactions on
  knowledge and data engineering 22~(10) (2009) 1345--1359.

\bibitem{skinner1965science}
B.~F. Skinner, Science and human behavior, Simon and Schuster, 1965.

\bibitem{celiberto2016transfer}
L.~A. Celiberto, R.~A. Bianchi, P.~E. Santos, Transfer learning heuristically
  accelerated algorithm: a case study with real robots, in: 2016 XIII Latin
  American Robotics Symposium and IV Brazilian Robotics Symposium (LARS/SBR),
  IEEE, 2016, pp. 311--316.

\bibitem{barrett2010transfer}
S.~Barrett, M.~E. Taylor, P.~Stone, Transfer learning for reinforcement
  learning on a physical robot, in: Ninth International Conference on
  Autonomous Agents and Multiagent Systems-Adaptive Learning Agents Workshop
  (AAMAS-ALA), Vol.~1, 2010, pp. 24--99.

\bibitem{parisotto2015actor}
E.~Parisotto, J.~L. Ba, R.~Salakhutdinov, Actor-mimic: Deep multitask and
  transfer reinforcement learning, arXiv preprint arXiv:1511.06342 (2015).

\bibitem{devin2017learning}
C.~Devin, A.~Gupta, T.~Darrell, P.~Abbeel, S.~Levine, Learning modular neural
  network policies for multi-task and multi-robot transfer, in: 2017 IEEE
  International Conference on Robotics and Automation (ICRA), IEEE, 2017, pp.
  2169--2176.

\bibitem{taylor2007cross}
M.~E. Taylor, P.~Stone, Cross-domain transfer for reinforcement learning, in:
  Proceedings of the 24th international conference on Machine learning, 2007,
  pp. 879--886.

\bibitem{konda2000actor}
V.~R. Konda, J.~N. Tsitsiklis, Actor-critic algorithms, in: Advances in neural
  information processing systems, 2000, pp. 1008--1014.

\bibitem{wiering2008ensemble}
M.~A. Wiering, H.~Van~Hasselt, Ensemble algorithms in reinforcement learning,
  IEEE Transactions on Systems, Man, and Cybernetics, Part B (Cybernetics)
  38~(4) (2008) 930--936.

\bibitem{sanchez2015priori}
E.~M. S{\'a}nchez, J.~B. Clempner, A.~S. Poznyak, A
  priori-knowledge/actor-critic reinforcement learning architecture for
  computing the mean--variance customer portfolio: the case of bank marketing
  campaigns, Engineering Applications of Artificial Intelligence 46 (2015)
  82--92.

\bibitem{lawhead2019bounded}
R.~J. Lawhead, A.~Gosavi, A bounded actor--critic reinforcement learning
  algorithm applied to airline revenue management, Engineering Applications of
  Artificial Intelligence 82 (2019) 252--262.

\bibitem{degris2012off}
T.~Degris, M.~White, R.~S. Sutton, Off-policy actor-critic, arXiv preprint
  arXiv:1205.4839 (2012).

\bibitem{mnih2016asynchronous}
V.~Mnih, A.~P. Badia, M.~Mirza, A.~Graves, T.~Lillicrap, T.~Harley, D.~Silver,
  K.~Kavukcuoglu, Asynchronous methods for deep reinforcement learning, in:
  International conference on machine learning, 2016, pp. 1928--1937.

\bibitem{silver2014deterministic}
D.~Silver, G.~Lever, N.~Heess, T.~Degris, D.~Wierstra, M.~Riedmiller,
  Deterministic policy gradient algorithms, in: Proceedings of the 31st
  International Conference on International Conference on Machine Learning,
  2014, pp. 1--9.

\bibitem{popov2017data}
I.~Popov, N.~Heess, T.~Lillicrap, R.~Hafner, G.~Barth-Maron, M.~Vecerik,
  T.~Lampe, Y.~Tassa, T.~Erez, M.~Riedmiller, Data-efficient deep reinforcement
  learning for dexterous manipulation, arXiv preprint arXiv:1704.03073 (2017).

\bibitem{ziebart2010modeling}
B.~D. Ziebart, Modeling purposeful adaptive behavior with the principle of
  maximum causal entropy, Machine Learning Department, School of Computer
  Science, Carnegie Mellon University (2010).

\bibitem{feldmann2012modularity}
S.~Feldmann, J.~Fuchs, B.~Vogel-Heuser, et~al., Modularity, variant and version
  management in plant automation--future challenges and state of the art, in:
  DS 70: Proceedings of DESIGN 2012, the 12th International Design Conference,
  Dubrovnik, Croatia, 2012, pp. 1689--1698.

\bibitem{eppinger2015product}
S.~Eppinger, K.~Ulrich, Product design and development, McGraw-Hill Higher
  Education, 2015.

\bibitem{simon1991architecture}
H.~A. Simon, The architecture of complexity, in: Facets of systems science,
  Springer, 1991, pp. 457--476.

\bibitem{gianetto2015network}
D.~A. Gianetto, B.~Heydari, Network modularity is essential for evolution of
  cooperation under uncertainty, Scientific reports 5 (2015) 9340.

\bibitem{mosleh2017fair}
M.~Mosleh, B.~Heydari, Fair topologies: Community structures and network hubs
  drive emergence of fairness norms, Scientific reports 7~(1) (2017) 1--9.

\bibitem{singh1992efficient}
S.~P. Singh, The efficient learning of multiple task sequences, in: Advances in
  neural information processing systems, 1992, pp. 251--258.

\bibitem{russell2003q}
S.~J. Russell, A.~Zimdars, Q-decomposition for reinforcement learning agents,
  in: Proceedings of the 20th International Conference on Machine Learning
  (ICML-03), 2003, pp. 656--663.

\bibitem{simpkins2019composable}
C.~Simpkins, C.~Isbell, Composable modular reinforcement learning, in:
  Proceedings of the AAAI Conference on Artificial Intelligence, Vol.~33, 2019,
  pp. 4975--4982.

\bibitem{andreas2016neural}
J.~Andreas, M.~Rohrbach, T.~Darrell, D.~Klein, Neural module networks, in:
  Proceedings of the IEEE conference on computer vision and pattern
  recognition, 2016, pp. 39--48.

\bibitem{chitnis2019learning}
R.~Chitnis, L.~P. Kaelbling, T.~Lozano-P{\'e}rez, Learning quickly to plan
  quickly using modular meta-learning, in: 2019 International Conference on
  Robotics and Automation (ICRA), IEEE, 2019, pp. 7865--7871.

\bibitem{harlow1949formation}
H.~F. Harlow, The formation of learning sets., Psychological review 56~(1)
  (1949) 51.

\bibitem{schweighofer2003meta}
N.~Schweighofer, K.~Doya, Meta-learning in reinforcement learning, Neural
  Networks 16~(1) (2003) 5--9.

\bibitem{schaul2010metalearning}
T.~Schaul, J.~Schmidhuber, Metalearning, Scholarpedia 5~(6) (2010) 4650.

\bibitem{lemke2015metalearning}
C.~Lemke, M.~Budka, B.~Gabrys, Metalearning: a survey of trends and
  technologies, Artificial intelligence review 44~(1) (2015) 117--130.

\bibitem{finn2017model}
C.~Finn, P.~Abbeel, S.~Levine, Model-agnostic meta-learning for fast adaptation
  of deep networks, in: Proceedings of the 34th International Conference on
  Machine Learning-Volume 70, JMLR. org, 2017, pp. 1126--1135.

\bibitem{grant2018recasting}
E.~Grant, C.~Finn, S.~Levine, T.~Darrell, T.~Griffiths, Recasting
  gradient-based meta-learning as hierarchical bayes, arXiv preprint
  arXiv:1801.08930 (2018).

\bibitem{finn2018probabilistic}
C.~Finn, K.~Xu, S.~Levine, Probabilistic model-agnostic meta-learning, in:
  Advances in Neural Information Processing Systems, 2018, pp. 9516--9527.

\bibitem{rakelly2019efficient}
K.~Rakelly, A.~Zhou, D.~Quillen, C.~Finn, S.~Levine, Efficient off-policy
  meta-reinforcement learning via probabilistic context variables, arXiv
  preprint arXiv:1903.08254 (2019).

\bibitem{gupta2018meta}
A.~Gupta, R.~Mendonca, Y.~Liu, P.~Abbeel, S.~Levine, Meta-reinforcement
  learning of structured exploration strategies, in: Advances in Neural
  Information Processing Systems, 2018, pp. 5302--5311.

\bibitem{schoettler2020meta}
G.~Schoettler, A.~Nair, J.~A. Ojea, S.~Levine, E.~Solowjow, Meta-reinforcement
  learning for robotic industrial insertion tasks, arXiv preprint
  arXiv:2004.14404 (2020).

\bibitem{todorov2012mujoco}
E.~Todorov, T.~Erez, Y.~Tassa, Mujoco: A physics engine for model-based
  control, in: 2012 IEEE/RSJ International Conference on Intelligent Robots and
  Systems, IEEE, 2012, pp. 5026--5033.

\bibitem{Mnih2015}
V.~Mnih, K.~Kavukcuoglu, D.~Silver, A.~A. Rusu, J.~Veness, M.~G. Bellemare,
  A.~Graves, M.~Riedmiller, A.~K. Fidjeland, G.~Ostrovski, S.~Petersen,
  C.~Beattie, A.~Sadik, I.~Antonoglou, H.~King, D.~Kumaran, D.~Wierstra,
  S.~Legg, D.~Hassabis, {Human-level control through deep reinforcement
  learning}, Nature 518~(7540) (2015) 529--533.
\newblock \href {https://doi.org/10.1038/nature14236}
  {\path{doi:10.1038/nature14236}}.

\bibitem{Sutton2000}
R.~S. Sutton, D.~McAllester, S.~Singh, Y.~Mansour, {Policy gradient methods for
  reinforcement learning with function approximation}, in: Advances in Neural
  Information Processing Systems, 2000, pp. 1038--1044.

\bibitem{mnih2013playing}
V.~Mnih, K.~Kavukcuoglu, D.~Silver, A.~Graves, I.~Antonoglou, D.~Wierstra,
  M.~Riedmiller, Playing atari with deep reinforcement learning, arXiv preprint
  arXiv:1312.5602 (2013).

\bibitem{kapturowski2018recurrent}
S.~Kapturowski, G.~Ostrovski, J.~Quan, R.~Munos, W.~Dabney, Recurrent
  experience replay in distributed reinforcement learning, in: International
  conference on learning representations, 2018, pp. 1--19.

\bibitem{van2016deep}
H.~Van~Hasselt, A.~Guez, D.~Silver, Deep reinforcement learning with double
  q-learning, in: Thirtieth AAAI Conference on Artificial Intelligence, 2016,
  pp. 1--13.

\bibitem{ziebart2008maximum}
B.~D. Ziebart, A.~L. Maas, J.~A. Bagnell, A.~K. Dey, Maximum entropy inverse
  reinforcement learning., in: Aaai, Vol.~8, Chicago, IL, USA, 2008, pp.
  1433--1438.

\bibitem{luo2016model}
B.~Luo, D.~Liu, T.~Huang, D.~Wang, Model-free optimal tracking control via
  critic-only q-learning, IEEE transactions on neural networks and learning
  systems 27~(10) (2016) 2134--2144.

\bibitem{yun2020evaluating}
A.~Yun, Evaluating the robustness of natural language reward shaping models to
  spatial relations, The University of Texas at Austin (2020).

\end{thebibliography}

\end{document}